\documentclass[10pt,twocolumn,letterpaper]{article}

\usepackage{iccv}
\usepackage{times}
\usepackage{epsfig}
\usepackage{graphicx}
\usepackage{amsmath}
\usepackage{amssymb}
\usepackage{lipsum}
\usepackage{balance}

\usepackage{enumitem}
\usepackage{import}
\usepackage{scalerel, stackengine}
\stackMath
\newcommand\reallywidehat[1]{%
\savestack{\tmpbox}{\stretchto{%
  \scaleto{%
    \scalerel*[\widthof{\ensuremath{#1}}]{\kern.1pt\mathchar"0362\kern.1pt}%
    {\rule{0ex}{\textheight}}
  }{\textheight}%
}{2.4ex}}%
\stackon[-6.9pt]{#1}{\tmpbox}%
}

\usepackage{bm}
\let\mathcal\undefined
\DeclareMathAlphabet{\mathcal}{OMS}{cmsy}{m}{n}

\DeclareSymbolFont{Xlargesymbols}{OMX}{cmex}{m}{n}
\DeclareMathSymbol{\Xsum}{\mathop}{Xlargesymbols}{80}

\usepackage{tabularx, booktabs}
\usepackage{multirow}
\newcolumntype{Y}{>{\centering\arraybackslash}X}

\usepackage{cuted} 
\usepackage{capt-of} 

\usepackage{placeins} 
\usepackage{flafter}  

\usepackage{mathtools}

\usepackage{ntheorem}
\theoremstyle{break}

\newtheorem{lemma}{Lemma}

\newcommand*{\QEDA}{\hfill\ensuremath{\blacksquare}}%

\DeclareMathOperator*{\argmin}{\arg\!\min}

\usepackage[pagebackref=true,breaklinks=true,letterpaper=true,colorlinks,bookmarks=false]{hyperref}

\iccvfinalcopy 

\def\iccvPaperID{****} 

\begin{document}

\title{
Closed-Form Optimal Two-View Triangulation Based on Angular Errors}

\author{Seong Hun Lee \hspace{25pt} Javier Civera \\
I3A, University of Zaragoza, Spain\\
{\tt\small \{seonghunlee, jcivera\}@unizar.es}
}
\maketitle


\begin{abstract}
In this paper, we study closed-form optimal solutions to two-view triangulation with known internal calibration and pose.
By formulating the triangulation problem as $L_1$ and $L_\infty$ minimization of angular reprojection errors,
we derive the exact closed-form solutions that guarantee global optimality under respective cost functions.
To the best of our knowledge, we are the first to present such solutions.
Since the angular error is rotationally invariant, our solutions can be applied for any type of central cameras, be it perspective, fisheye or omnidirectional. 
Our methods also require significantly less computation than the existing optimal methods.
Experimental results on synthetic and real datasets validate our theoretical derivations.
\end{abstract}

\section{Introduction}
\label{sec:intro}
Recovering the position of a 3D point given its projections in two or more cameras is called triangulation.
It constitutes a fundamental building block in stereo vision \cite{tippetts2016review}, simultaneous localization and mapping (SLAM) \cite{orb-slam2} and structure-from-motion (SfM) pipelines \cite{sfm_revisited}.
For large problems, reconstructing thousands (or millions) of points is not uncommon, 
so achieving fast and accurate triangulation is important for the performance of such systems.



If one assumes the exact knowledge of camera matrices and noiseless feature measurements, triangulation amounts to intersecting two backprojected rays that correspond to the same 3D point.
In practice, however, this assumption is unrealistic, and the rays do not necessarily intersect. 
Therefore, a nontrivial method is required even for just two views.

The standard approach is to find the 3D point that minimizes a chosen cost function given the feature measurements. 
The most common are the $L_1$ norm (sum of magnitude), $L_2$ norm (sum of squares) and $L_\infty$ norm (maximum) of image reprojection errors.
While reasonable for perspective cameras, the image reprojection error does not generalize well to different camera types (e.g., omnidirectional or fully spherical panoramic cameras).
This motivates the use of \textit{angular} reprojection error, a rotationally invariant alternative to the \textit{image} reprojection error that is generic and independent of the projection geometry \cite{mouragnon2009generic,closed_form_oliensis,pagani2011structure}.

In this work, we derive, for the first time to our knowledge, the exact closed-form solutions to the $L_1$ and $L_\infty$ optimal triangulation from two views based on the angular reprojection error.
Unlike iterative methods (e.g., \cite{kanatani, lindstrom}), the proposed methods guarantee global optimality without any iterations, and unlike polynomial methods (e.g., \cite{hartley_triangulation, nister_phd_thesis, stewenius_three_view}), they do not involve finding the roots of a higher-degree polynomial.
Hence, our methods simultaneously provide the global optimality, speed and simplicity. 
We also present our own derivation of the $L_2$ optimal solution that is much more compact and geometrically intuitive than the existing one \cite{closed_form_oliensis}.
Since all three methods are based on the angular error, they are not limited to standard perspective cameras and can also be used for fisheye, omnidirectional and fully spherical panoramic cameras.


The paper is organized as follows.
In the next three sections, we discuss the related work and preliminaries.
Section \ref{sec:L_1}, \ref{sec:L_2} and \ref{sec:L_inf} respectively present the closed-form solutions to the $L_1$, $L_2$ and $L_\infty$ optimal triangulation.
To make our paper compact and easily accessible, we separated the proofs from our main findings and put them in the appendix.
Section \ref{sec:cheirality} addresses the cheirality constraint.
Finally, experimental results are provided in Section \ref{sec:results}, followed by the conclusions in Section \ref{sec:conclusion}. 


\section{Related Work}
\label{sec:related_work}
The most widespread approach to triangulation is to find the 3D point that minimizes the $L_2$ norm of image reprojection errors \cite{hartley_book}.
Assuming that image points are perturbed by Gaussian noise, the $L_2$ optimal solution gives the maximum likelihood estimate (MLE).
This can be obtained in closed form by solving a polynomial of degree 6 for two views \cite{hartley_triangulation} and degree 47 for three views \cite{stewenius_three_view}.
Such polynomial methods are, however, computationally expensive and susceptible to ill-conditioning \cite{lindstrom}.
Besides, an iterative search for the roots may converge to a local minimum \cite{hartley_triangulation}.

Another two-view method by Kanatani et al. \cite{kanatani}  iteratively corrects the 2D projections of the points.
Although this method was shown to be faster than the one by Hartley and Sturm \cite{hartley_triangulation}, it does not satisfy the epipolar constraint \cite{eight_point} in each iteration. 
Lindstrom \cite{lindstrom} solved this problem with an improved iterative algorithm that is even more stable and faster.   
However, neither his method nor Kanatani's guarantees global optimality.
Oliensis \cite{closed_form_oliensis} showed that by formulating the problem as $L_2$ minimization of the sine of angular reprojection errors, an exact closed-form solution can be derived for two-view triangulation.

Instead of minimizing the $L_2$ norm, one may choose to minimize the $L_1$ norm of reprojection errors.
The advantage of $L_1$ norm is that it is more robust to outliers as it places less emphasis on large errors \cite{hartley_triangulation, L2_kahl}. 
For two views, Hartley and Sturm \cite{hartley_triangulation} showed that the $L_1$ optimal solution can be obtained in closed form by solving a polynomial of degree 8.
They also found that the $L_1$ optimization gives slightly more accurate 3D results than the $L_2$ optimization.

In geometric problems, another popular norm is the $L_\infty$ norm.
The $L_\infty$ optimal solution corresponds to the MLE under the assumption of uniform noise in the image points \cite{optimal_algorithms_hartley}. 
The advantage of the $L_\infty$ cost function over the $L_2$ cost is that it is relatively simpler and has a single minimum \cite{L_inf_hartley}. 
For the case of two views, N\'{i}ster \cite{nister_phd_thesis} showed that the optimal solution can be obtained in closed form by keeping the reprojection errors equal in the two views and solving the resulting quartic equation. 
A main drawback of the $L_\infty$ cost is that it is relatively more sensitive to outliers \cite{L_inf_hartley}.
This being said, such sensitivity was shown to be useful for outlier removal \cite{L_inf_outlier_removal_sim, L_inf_outlier_removal_olsson, L_inf_outlier_removal_li}.

While most of the aforementioned works formulate their optimization problem in terms of the image reprojection error, the angular reprojection error is another popular choice.
It embodies a better noise model for fisheye or omnidirectional cameras \cite{closed_form_oliensis, sfm_wideangle}.
Even for perspective cameras, the assumption of Gaussian noise is not justified \cite{optimal_algorithms_hartley}, and the angular reprojection error is just as valid as the image reprojection error, if not more so.
In the literature, it has been proposed to minimize the sine of angular reprojection errors in $L_2$ norm \cite{closed_form_oliensis}, the tangent in $L_2$ or $L_\infty$ norm \cite{L2_verify_hartley,L_inf_hartley, L_inf_kahl}, and the cosine in negative $L_1$ norm \cite{recker_multiview_angular_error, chesi_multiview_angular_error}. 
In contrast to these methods, our $L_1$ and $L_\infty$ optimization do not involve trigonometric functions.

\section{Preliminaries on 3D Geometry}
\label{sec:prelim_on_geometry}
Throughout the paper, we adopt the following notation:
We use bold letters for vectors and matrices, and light letters for scalars.
The Euclidean norm of a vector $\mathbf{v}$ is denoted by $\lVert \mathbf{v} \rVert$, and the unit vector by $\widehat{\mathbf{v}}={\mathbf{v}}/{\lVert\mathbf{v}\rVert}$.
The angle between two lines $\mathbf{L}_0$ and $\mathbf{L}_1$ is denoted by $\angle\left(\mathbf{L}_0, \ \mathbf{L}_1\right)\in[0, \pi/2]$.

The following vector identities will come in handy later:
\vspace{-0.8em}
\begin{equation}
\label{eq:triple_product_invariance}
    \mathbf{a}\cdot (\mathbf{b}\times\mathbf{c})=\mathbf{b}\cdot (\mathbf{c}\times\mathbf{a})=\mathbf{c}\cdot (\mathbf{a}\times\mathbf{b}) 
    \vspace{-0.5em}
\end{equation}
\vspace{-0.5em}
\begin{equation}
\label{eq:|axb|^2}
    \lVert\widehat{\mathbf{a}}\times\widehat{\mathbf{b}}\rVert^2=1-(\widehat{\mathbf{a}}\cdot\widehat{\mathbf{b}})^2
\end{equation} 
We also make frequent use of the following formulas:
\begin{enumerate}[leftmargin=*]\itemsep0em
\vspace{-0.2em}
    \item The distance between a point $\mathbf{p}$ and a plane $\Pi_0(\mathbf{x})=\mathbf{n}_0\cdot(\mathbf{x}-\mathbf{c}_0)=0$ is given by $\lVert\mathbf{p}-\mathbf{r}_0\rVert$ where $\mathbf{r}_0$ is the projection of $\mathbf{p}$ onto $\Pi_0$.
    This is computed as follows:
    \vspace{-0.5em}
    \begin{equation}
    \label{eq:shortest_distance_point_plane}
        \lVert\mathbf{p}-\mathbf{r}_0\rVert 
        =
        |\widehat{\mathbf{n}}_0\cdot(\mathbf{p}-\mathbf{c}_0)|.
        \vspace{-0.5em}
    \end{equation}
    
    \item The distance between two skew lines $\mathbf{L}_0(s_0)=\mathbf{c}_0+s_0{\mathbf{m}}_0$ and $\mathbf{L}_1(s_1)=\mathbf{c}_1+s_1{\mathbf{m}}_1$ is given by  $\lVert\mathbf{r}_0-\mathbf{r}_1\rVert$ where $\mathbf{r}_0$ and $\mathbf{r}_1$ are the points on each line that form the closest pair.
    Letting $\mathbf{t} = \mathbf{c}_0-\mathbf{c}_1$ and $\mathbf{q}=\mathbf{m}_0\times\mathbf{m}_1$, this is computed as follows:
    \vspace{-0.5em}
    \begin{equation}
    \label{eq:shortest_distance_skew_rays}
        \lVert\mathbf{r}_0-\mathbf{r}_1\rVert = \left|\mathbf{t}\cdot\widehat{\mathbf{q}}\right|.
        \vspace{-0.5em}
    \end{equation}
    The two points can also be obtained individually \cite{kanazawa}:
    \vspace{-0.5em}
    \begin{align}
        \mathbf{r}_0 
        &=
        \mathbf{c}_0+\frac{\mathbf{q}\cdot\left({\mathbf{m}}_1\times\mathbf{t}\right)}{\lVert\mathbf{q}\rVert^{2}}{\mathbf{m}}_0, \label{eq:closest_ponits1}\\
        \mathbf{r}_1 
        &=
        \mathbf{c}_1+\frac{\mathbf{q}\cdot\left({\mathbf{m}}_0\times\mathbf{t}\right)}{\lVert\mathbf{q}\rVert^{2}}{\mathbf{m}}_1.
        \label{eq:closest_ponits2}
    \end{align}
\end{enumerate}\vspace{-0.5em}
Equation \eqref{eq:shortest_distance_skew_rays} can be interpreted as the minimum amount of translation required for the two lines to intersect.
In this work, it will be also important to know the minimum amount of rotation (or pivot) required for the two lines to intersect.
We answer this question in the following lemma:

\begin{lemma}[Minimum Pivot Angle for Intersection]
\label{lemma:single_pivot}
Given two skew lines ${\mathbf{L}_0(s_0) = \mathbf{c}_0+s_0{\mathbf{m}}_0}$ and $\mathbf{L}_1(s_1) = \mathbf{c}_1+s_1{\mathbf{m}}_1$, 
let $\mathbf{L}_0'$ be the line that forms the smallest angle $\theta_0\in[0, \ \pi/2]$ to $\mathbf{L}_0$ among all possible lines that intersect both point $\mathbf{c}_0$ and line $\mathbf{L}_1$.
Then, $\mathbf{L}'_0$ is the projection of $\mathbf{L}_0$ onto the plane that contains $\mathbf{c}_0$ and $\mathbf{L}_1$. 
Furthermore, letting $\mathbf{t}=\mathbf{c}_0-\mathbf{c}_1$ and $\mathbf{n}_1 = \mathbf{m}_1\times\mathbf{t}$, 
\vspace{-0.5em}
\begin{equation}
\label{eq:lemma_single_pivot:1}
\sin{(\theta_0)} 
= 
|\widehat{\mathbf{n}}_1\cdot\widehat{\mathbf{m}}_0|.
\vspace{-0.5em}
\end{equation}
We call $\theta_0$ \textbf{the minimum pivot angle for intersection}, as it represents the smallest angle required for pivoting line $\mathbf{L}_0$ at $\mathbf{c}_0$ to make it intersect $\mathbf{L}_1$.
\end{lemma}
\vspace{-0.5em}
\noindent\textbf{Proof. }  
Refer to Appendix \ref{app:proof_single_pivot}. \QEDA

\section{Preliminaries on Two-View Triangulation}
\label{sec:prelim_on_triangulation}
\begin{figure}[t]
 \centering
 \includegraphics[width=0.27\textwidth]{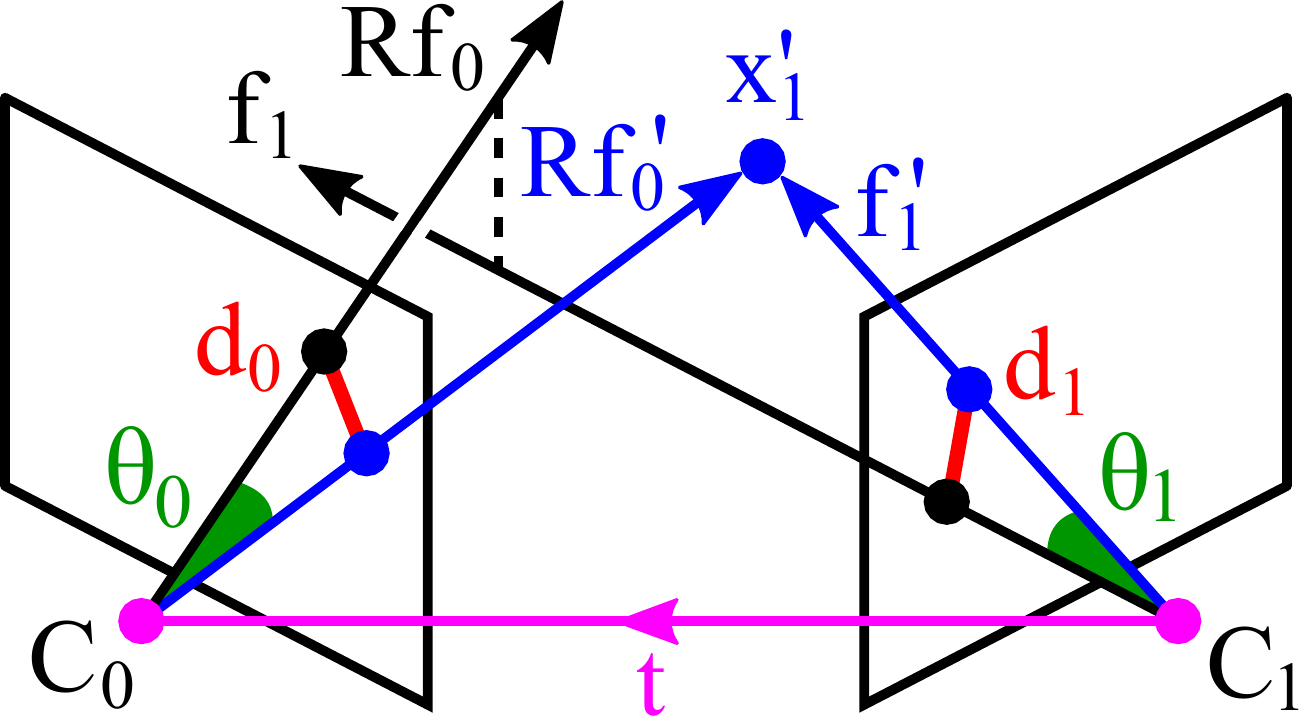}
\caption{The difference between the observed features ($\mathbf{f}_0$, $\mathbf{f}_1$) and the triangulation result ($\mathbf{f}'_0$, $\mathbf{f}'_1$) can be quantified by either image reprojection errors ($d_0$, $d_1$) or angular reprojection errors ($\theta_0$, $\theta_1$). 
 }
\label{fig:reprojection_error}
\end{figure}

Consider two cameras $C_0$ and $C_1$ observing the same 3D world point $\mathbf{x}_w$. 
Let $\mathbf{c}_0$ and $\mathbf{c}_1$ be their positions in the world frame, and let $\mathbf{R}$ and $\mathbf{t}$ be the rotation matrix and translation vector that together transform a point from the camera frame $C_0$ to $C_1$, i.e., $\mathbf{x}_1=\mathbf{R}\mathbf{x}_0+\mathbf{t}$, where $\mathbf{x}_0 = [x_0, y_0, z_0]^\intercal$ and $\mathbf{x}_1 = [x_1, y_1, z_1]^\intercal$ correspond to $\mathbf{x}_w$ in camera frame $C_0$ and $C_1$, respectively.
Since triangulation is impossible for zero translation, we set $\left\Vert \mathbf{t}\right\Vert =\left\Vert \mathbf{c}_0-\mathbf{c}_1\right\Vert= 1$ without loss of generality.
Let $\mathbf{u}_0 = (u_0, v_0,1)^\intercal$ and $\mathbf{u}_1=(u_1, v_1,1)^\intercal$ be the homogeneous pixel coordinates of the estimated correspondence to $\mathbf{x}_w$ in each frame.
Given the camera calibration matrix $\mathbf{K}$, the normalized image coordinates $\mathbf{f}_0=[x_0/z_0, y_0/z_0, 1]^\intercal$ and $\mathbf{f}_1=[x_1/z_1, y_1/z_1, 1]^\intercal$ are related to $\mathbf{u}_0$ and $\mathbf{u}_1$ by
$\mathbf{u}_0 = \mathbf{K}\mathbf{f}_0$ and $\mathbf{u}_1 = \mathbf{K}\mathbf{f}_1$. 

The two backprojected rays in frame $C_1$, i.e.,  $\mathbf{r}_1(s_1) = s_1\mathbf{f}_1$ and $\mathbf{r}_0(s_0)=s_0\mathbf{Rf}_0+\mathbf{t}$, do not necessarily intersect due to inaccuracies in the image measurements and camera matrices.
For the rays to intersect, $\mathbf{f}_0$ and $\mathbf{f}_1$ must be corrected to $\mathbf{f}'_0$ and $\mathbf{f}'_1$ such that the epipolar constraint \cite{eight_point} is satisfied. 
It enforces the coplanarity of $\mathbf{f}'_1$, $\mathbf{Rf}'_0$ and $\mathbf{t}$, and is given by
\begin{equation}
    \label{eq:epipolar_constraint}
    {\mathbf{f}'_1}\cdot\left(\mathbf{t}\times\mathbf{R}\mathbf{f}'_0\right)=0.
\end{equation}

The goal of the optimal triangulation is to \textit{minimally} correct the feature rays so that they satisfy \eqref{eq:epipolar_constraint} and intersect at some point $\mathbf{x}'_1$ in frame $C_1$.
What is meant by ``minimal" depends on the chosen cost function and error criterion.
Fig. \ref{fig:reprojection_error} illustrates two most popular error criteria, namely the image reprojection error and the angular reprojection error.
Formally, they are defined as follows:
\vspace{-0.1em}
\begin{flalign}
    &d_i := \lVert\mathbf{u}_i-\mathbf{u}'_i\rVert =\lVert\mathbf{K}\left(\mathbf{f}_i-\mathbf{f}'_i\right)\rVert, & \text{for} \ \  i=0,1 \label{eq:img_reproj-error}\\
    &\theta_i := \angle\left(\mathbf{f}_i, \mathbf{f}'_i\right)=\angle\left(\mathbf{K}^{-1}\mathbf{u}_i, \mathbf{K}^{-1}\mathbf{u}'_i\right) & \text{for} \ \ i=0,1 \label{eq:ang_reproj_error}
\end{flalign}
\vspace{-1.2em}

In this work, we minimize the latter in $L_1$, $L_2$ and $L_\infty$ norms.
Once we have the optimal $\mathbf{f}'_0$ and $\mathbf{f}'_1$, the point of intersection $\mathbf{x}'_1$ can be obtained using either \eqref{eq:closest_ponits1} or \eqref{eq:closest_ponits2}:
\vspace{-0.1em}
\begin{gather}
\label{eq:3d_point}
    \mathbf{x}'_1= \mathbf{t}+\underbrace{\frac{\mathbf{z}\cdot\left(\mathbf{t}\times\mathbf{f}'_1\right)}{\lVert\mathbf{z}\rVert^2}}_{\lambda_0}\mathbf{Rf}'_0 
    =
    \underbrace{\frac{\mathbf{z}\cdot\left(\mathbf{t}\times\mathbf{R}\mathbf{f}'_0\right)}{\lVert\mathbf{z}\rVert^2}}_{\lambda_1}\mathbf{f}'_1 \hspace{-1em} \\
    \text{with} \quad \mathbf{z}=\mathbf{f}'_1\times\mathbf{R}\mathbf{f}'_0, \nonumber
\end{gather}
\vspace{-1.5em}

\noindent where $\lambda_i$ equals the depth multiplied by $\lVert\mathbf{f}'_i\rVert$ for $i=0,1$.

\begin{figure}[t]
 \centering
 \includegraphics[width=0.43\textwidth]{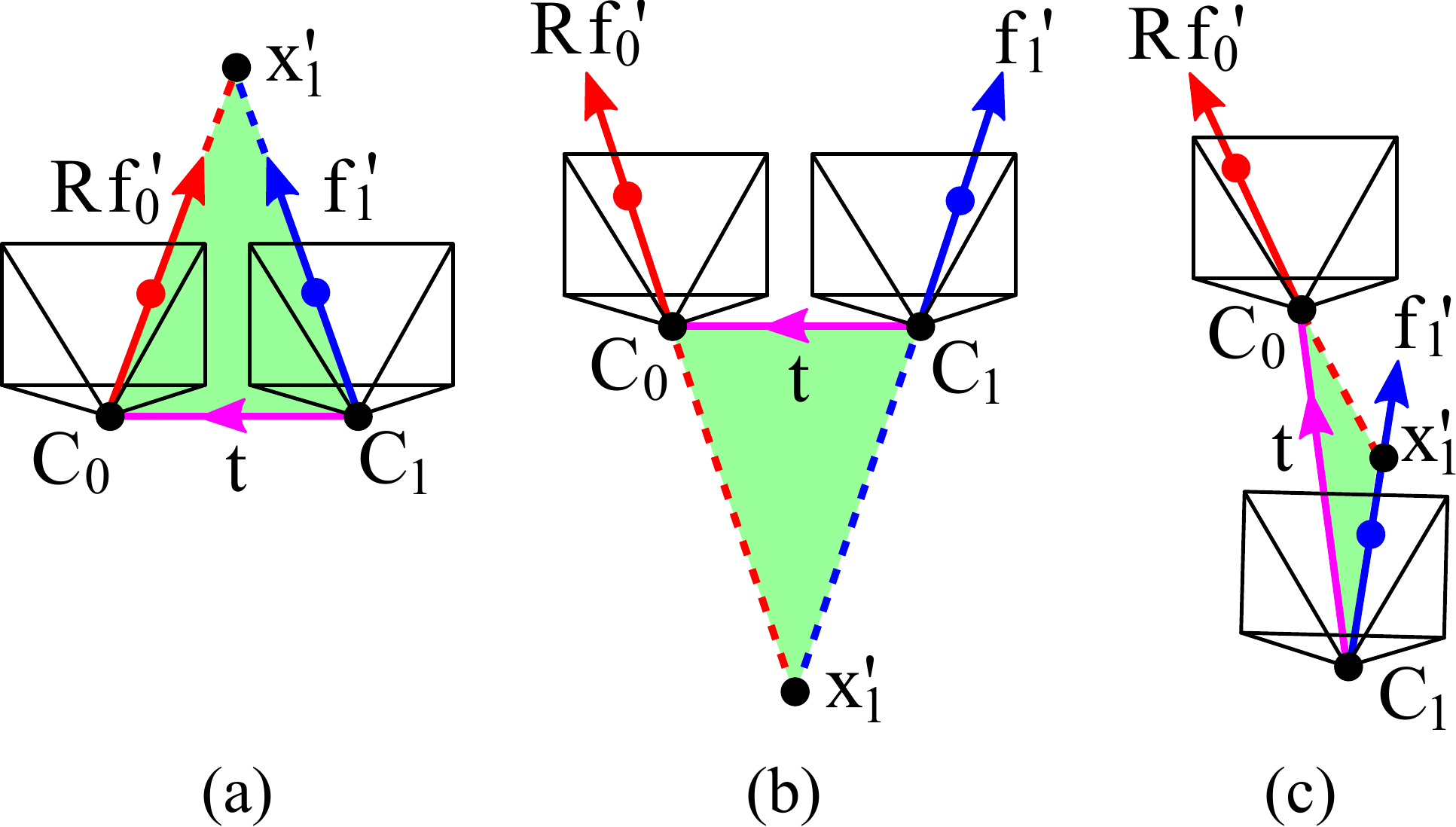}
\caption{Example scenarios satisfying the epipolar constraint \eqref{eq:epipolar_constraint}.
The epipolar plane (shown in green) contains both the rays and the camera centers. Cheirality condition is violated in case (b) and (c).}
\label{fig:epipolar_constraint}
\end{figure}

Note that the epipolar constraint \eqref{eq:epipolar_constraint} is a necessary condition for intersecting the two rays, but not a sufficient one.
Fig. \ref{fig:epipolar_constraint} illustrates scenarios where the two rays are coplanar, but do not intersect.
This happens when the intersection requires negative depth(s), violating the \textit{cheirality} constraint \cite{hartley_book}.
In the following analysis (until Section \ref{sec:cheirality}), we will temporarily assume that satisfying the epipolar constraint \eqref{eq:epipolar_constraint} is sufficient for intersecting the rays. 

\section{Closed-Form $L_1$ Triangulation}
\label{sec:L_1}
The $L_1$ triangulation based on the angular reprojection error \eqref{eq:ang_reproj_error} finds the feature rays $\mathbf{f}'_0$ and $\mathbf{f}'_1$ that minimize $\theta_0+\theta_1$ subject to the epipolar constraint \eqref{eq:epipolar_constraint}.
The following lemma reveals a surprising fact that $(\theta_0+\theta_1)_\text{min}$ is achieved by correcting either one of $\mathbf{f}_0$ or $\mathbf{f}_1$, but not both:

\break
\begin{lemma}[$L_1$ Angle Minimization]
\label{lemma:L_1}
Given two skew lines $\mathbf{L}_0(s_0) = \mathbf{c}_0+s_0{\mathbf{m}}_0$ and $\mathbf{L}_1(s_1) = \mathbf{c}_1+s_1{\mathbf{m}}_1$, consider any two intersecting lines that also pass $\mathbf{c}_0$ and $\mathbf{c}_1$, respectively, i.e.,  $\mathbf{L}_0'(s_0')=\mathbf{c}_0+s_0'{\mathbf{m}}_0'$ and $\mathbf{L}_1'(s_1')=\mathbf{c}_1+s_1'{\mathbf{m}}_1'$. 
Let $\mathbf{t}=\mathbf{c}_0-\mathbf{c}_1$, $\mathbf{n}_0=\mathbf{m}_0\times\mathbf{t}$, $\mathbf{n}_1=\mathbf{m}_1\times\mathbf{t}$, $\theta_0=\angle(\mathbf{L}_0, \mathbf{L}_0')$ and $\theta_1=\angle(\mathbf{L}_1, \mathbf{L}_1')$.
Then, $(\theta_0+\theta_1)$ is minimized for the following $\mathbf{m}'_0$ and $\mathbf{m}'_1$:
\vspace{-0.5em}
\noindent
\begin{flalign}
    &\text{If} \quad \lVert\widehat{\mathbf{m}}_0\times\mathbf{t}\rVert \leq \lVert\widehat{\mathbf{m}}_1\times\mathbf{t}\rVert, \nonumber&&\\
    &\quad\mathbf{m}'_0 = \mathbf{m}_0-\left(\mathbf{m}_0\cdot\widehat{\mathbf{n}}_1\right)\widehat{\mathbf{n}}_1 \quad\text{and}\quad \mathbf{m}'_1 = \mathbf{m}_1.  \label{eq:L_1:1}\\
    &\text{Otherwise},\nonumber&&\\
    &\quad\mathbf{m}'_0 = \mathbf{m}_0 \quad\text{and}\quad \quad\mathbf{m}'_1 = \mathbf{m}_1-\left(\mathbf{m}_1\cdot\widehat{\mathbf{n}}_0\right)\widehat{\mathbf{n}}_0 \label{eq:L_1:2}
\end{flalign}
\end{lemma}
\vspace{-0.5em}
\noindent\textbf{Proof. }  
Refer to Appendix \ref{app:proof_L_1}. \QEDA

By substituting $\mathbf{Rf}_0$ and $\mathbf{f}_1$ into $\mathbf{m}_0$ and $\mathbf{m}_1$ in the above lemma, the resulting $\mathbf{m}'_0$ and $\mathbf{m}'_1$ become the corrected rays $\mathbf{Rf}'_0$ and $\mathbf{f}'_1$ that satisfy the $L_1$ optimality, and $\mathbf{n}_0$ (or $\mathbf{n}_1$) becomes the normal of the corresponding epipolar plane.

\section{Closed-Form $L_2$ Triangulation}
\label{sec:L_2}
Considering that the angular errors are small in practice, the ``relaxed" $L_2$ triangulation finds the feature rays $\mathbf{f}'_0$ and $\mathbf{f}'_1$ that minimize
$\sin^2(\theta_0)+\sin^2(\theta_1)$ (instead of $\theta_0^2+\theta_1^2$) subject to the epipolar constraint \eqref{eq:epipolar_constraint}.
Note that the small-angle approximation by $\sin(\theta)$ is more accurate than by $\tan(\theta)$ or $1-\cos(\theta)$ that have been used in literature \cite{L2_verify_hartley, L_inf_hartley, L_inf_kahl, recker_multiview_angular_error, chesi_multiview_angular_error}.
This is easily seen by comparing their Maclaurin expansions.
As will be shown in the following lemma (and previously in \cite{closed_form_oliensis}), the relaxation with the sine function allows us to derive the $L_2$ optimal solution in closed form.

\begin{lemma}[$L_2$ Angle Minimization]
\label{lemma:L_2}
Given two skew lines $\mathbf{L}_0(s_0) = \mathbf{c}_0+s_0{\mathbf{m}}_0$ and $\mathbf{L}_1(s_1) = \mathbf{c}_1+s_1{\mathbf{m}}_1$, consider any two intersecting lines that also pass $\mathbf{c}_0$ and $\mathbf{c}_1$, respectively, i.e.,  $\mathbf{L}_0'(s_0')=\mathbf{c}_0+s_0'{\mathbf{m}}_0'$ and $\mathbf{L}_1'(s_1')=\mathbf{c}_1+s_1'{\mathbf{m}}_1'$. 
Let $\mathbf{t}=\mathbf{c}_0-\mathbf{c}_1$,  $\theta_0=\angle(\mathbf{L}_0, \mathbf{L}_0')$ and $\theta_1=\angle(\mathbf{L}_1, \mathbf{L}_1')$.
Then, $\left(\sin^2{\theta_0}+\sin^2{\theta_1}\right)$ is minimized for
\vspace{-0.5em}
\begin{equation}
\label{eq:lemma:L_2:1}
    {\mathbf{m}}_i' 
    =
    \mathbf{m}_i - (\mathbf{m}_i\cdot \widehat{\mathbf{n}}')\widehat{\mathbf{n}}' \quad \text{for} \quad i=0,1,
\end{equation}
where $\widehat{\mathbf{n}}'$ is the second column of the $3\times3$ matrix $\mathbf{V}$ from
\vspace{-0.5em}
\begin{equation}
\label{eq:lemma:L_2:2}
    \mathbf{USV}\strut^\intercal = \text{SVD}\left(\begin{bmatrix}\widehat{\mathbf{m}}_0 \quad \widehat{\mathbf{m}}_1
    \end{bmatrix}^\intercal\left(\mathbf{I}-\widehat{\mathbf{t}}\ {\widehat{\mathbf{t}}} \hspace{1pt} \strut^\intercal\right)\right). 
    \vspace{-0.5em}
\end{equation}
\end{lemma}
\vspace{-0.5em}
\noindent\textbf{Proof. }  
Refer to Appendix \ref{app:proof_L_2}. \QEDA

Analogously to the $L_1$ method, substituting $\mathbf{Rf}_0$ and $\mathbf{f}_1$ into $\mathbf{m}_0$ and $\mathbf{m}_1$ in the above lemma gives $\mathbf{Rf}'_0=\mathbf{m}'_0$ and $\mathbf{f}'_1 =\mathbf{m}'_1$ that satisfy the $L_2$ optimality. 

{\renewcommand{\arraystretch}{1.1}%
\begin{table}[t]
\begin{center}
\begin{tabular}{|p{0.1em}p{0.1em}p{0.38\textwidth}|}
\hline
\textbf{Input:} &\multicolumn{2}{l|}{\ \ \ \ \ \ \  Calib. matrix ($\mathbf{K}$), relative pose ($\mathbf{R}$, $\mathbf{t}$), and a} \\
&\multicolumn{2}{l|}{\ \ \ \ \ \ \ match ($\mathbf{u}_0$, $\mathbf{u}_1$) from two views ($C_0, C_1$).}\\
\textbf{Output:} &\multicolumn{2}{l|}{\ \ \ \ \ \ \ \ \ \  Triangulated 3D point ($\mathbf{x}'_1$) in ref. frame $C_1$.}\\
1) & \multicolumn{2}{l|}{$\mathbf{f}_0\leftarrow\mathbf{K}^{{\text{--1}}}\mathbf{u}_0$, $\mathbf{f}_1\leftarrow\mathbf{K}^{{\text{--1}}}\mathbf{u}_1$, $\mathbf{m}_0\leftarrow\mathbf{Rf}_0$,  $\mathbf{m}_1\leftarrow\mathbf{f}_1$.}\\
2)& 
\multicolumn{2}{l|}{\textbf{For} $\bm{L_1}$ \textbf{triangulation:}}
\\
& &
If $\lVert\widehat{\mathbf{m}}_0\times\mathbf{t}\rVert \leq \lVert\widehat{\mathbf{m}}_1\times\mathbf{t}\rVert$, use \eqref{eq:L_1:1} to obtain $\mathbf{m}'_0$ and $\mathbf{m}'_1$.
Otherwise, use \eqref{eq:L_1:2}.
\\
& \multicolumn{2}{l|}{\textbf{For} $\bm{L_2}$ \textbf{triangulation:}}
\\
& &
Compute $\mathbf{m}'_0$ and $\mathbf{m}'_1$ from \eqref{eq:lemma:L_2:1} and \eqref{eq:lemma:L_2:2}.
\\
& \multicolumn{2}{l|}{\textbf{For} $\bm{L_\infty}$ \textbf{triangulation:}}
\\
& &
Compute $\mathbf{m}'_0$ and $\mathbf{m}'_1$ from \eqref{eq:lemma:L_inf:1} and \eqref{eq:lemma:L_inf:2}.
\\
3)& 
\multicolumn{2}{p{0.4\textwidth}|}{$\mathbf{Rf}'_0\leftarrow\mathbf{m}'_0$ and  $\mathbf{f}'_1\leftarrow\mathbf{m}'_1$.}
\\
4)& 
\multicolumn{2}{p{0.4\textwidth}|}{Check cheirality:}
\\
& &
(i) \ Obtain $\lambda_0$ and $\lambda_1$ from \eqref{eq:3d_point}.

(ii) Discard the point and terminate if either

\hspace{13pt} $\lambda_0 \leq 0$ or $\lambda_1 \leq 0$.
\\
5)& 
\multicolumn{2}{p{0.4\textwidth}|}{Check angular reprojection errors:}
\\
& & 
(i) \ $\theta_0 \leftarrow \angle(\mathbf{Rf}_0, \mathbf{Rf}'_0)$ and  $\theta_1 \leftarrow \angle(\mathbf{f}_1, \mathbf{f}'_1)$.

(ii) Discard the point and terminate if 

\hspace{13pt} $\max(\theta_0, \theta_1) > \epsilon_1$ for some small  $\epsilon_1$.
\\
6)& 
\multicolumn{2}{p{0.4\textwidth}|}{Check parallax:}
\\
& & 
(i) \ $\beta \leftarrow \angle(\mathbf{Rf}'_0, \mathbf{f}'_1)$

(ii) Discard the point and terminate if 

\hspace{13pt} $\beta<\epsilon_2$ for some small $\epsilon_2$.
\\
7)& 
\multicolumn{2}{p{0.4\textwidth}|}{Compute and return $\mathbf{x}'_1$ from \eqref{eq:3d_point}.}
\\
\hline
\end{tabular}
\end{center}\vspace{-0.3em}
 \caption{Summary of the proposed methods.}
 \label{tab:summary}
\end{table}
}

\section{Closed-Form $L_\infty$ Triangulation}
\label{sec:L_inf}
The $L_\infty$ triangulation based on the angular reprojection error \eqref{eq:ang_reproj_error} finds the feature rays $\mathbf{f}'_0$ and $\mathbf{f}'_1$ that minimize $\max(\theta_0,\theta_1)$ subject to the epipolar constraint \eqref{eq:epipolar_constraint}.
The following lemma states that this is achieved when $\theta_0=\theta_1$:

\begin{lemma}[$L_\infty$ Angle Minimization]
\label{lemma:L_inf}
Given two skew lines $\mathbf{L}_0(s_0) = \mathbf{c}_0+s_0\mathbf{m}_0$ and $\mathbf{L}_1(s_1) = \mathbf{c}_1+s_1\mathbf{m}_1$, 
consider any two intersecting lines that also pass $\mathbf{c}_0$ and $\mathbf{c}_1$, respectively, i.e.,  $\mathbf{L}_0'(s_0')=\mathbf{c}_0+s_0'\mathbf{m}_0'$ and $\mathbf{L}_1'(s_1')=\mathbf{c}_1+s_1'\mathbf{m}_1'$.
Let $\mathbf{t}=\mathbf{c}_0-\mathbf{c}_1$,
$\mathbf{n}_a=\left(\widehat{\mathbf{m}}_0+\widehat{\mathbf{m}}_1\right)\times\mathbf{t}$,
$\mathbf{n}_b=\left(\widehat{\mathbf{m}}_0-\widehat{\mathbf{m}}_1\right)\times\mathbf{t}$,
$\theta_0=\angle(\mathbf{L}_0, \mathbf{L}_0')$ and $\theta_1=\angle(\mathbf{L}_1, \mathbf{L}_1')$.
Then, $\max{\left(\theta_0, \theta_1\right)}$ is minimized when $\theta_0=\theta_1$.
This is achieved for
\vspace{-0.5em}
\begin{equation}
\label{eq:lemma:L_inf:1}
    {\mathbf{m}}_i' 
    =
    \mathbf{m}_i - (\mathbf{m}_i\cdot \widehat{\mathbf{n}}')\widehat{\mathbf{n}}' \quad \text{for} \quad i=0,1,
\end{equation}
where 
\vspace{-1em}
\begin{equation}
    \mathbf{n}' = 
    \begin{cases}
    \mathbf{n}_a \ \ \text{ if } \  \lVert\mathbf{n}_a\rVert \geq \lVert\mathbf{n}_b\rVert \\
    \mathbf{n}_b \ \ \text{ ohterwise}\\
    \end{cases}
    \label{eq:lemma:L_inf:2}
\end{equation}
\end{lemma}
\vspace{-0.5em}
\noindent\textbf{Proof. }  
Refer to Appendix \ref{app:proof_L_inf}. \QEDA

Analogously to the previous two methods, substituting $\mathbf{Rf}_0$ and $\mathbf{f}_1$ into $\mathbf{m}_0$ and $\mathbf{m}_1$ gives $\mathbf{Rf}'_0=\mathbf{m}'_0$ and $\mathbf{f}'_1 =\mathbf{m}'_1$ that satisfy the $L_\infty$ optimality. 

\begin{figure*}[t]
\centering
\includegraphics[width=0.95\textwidth]{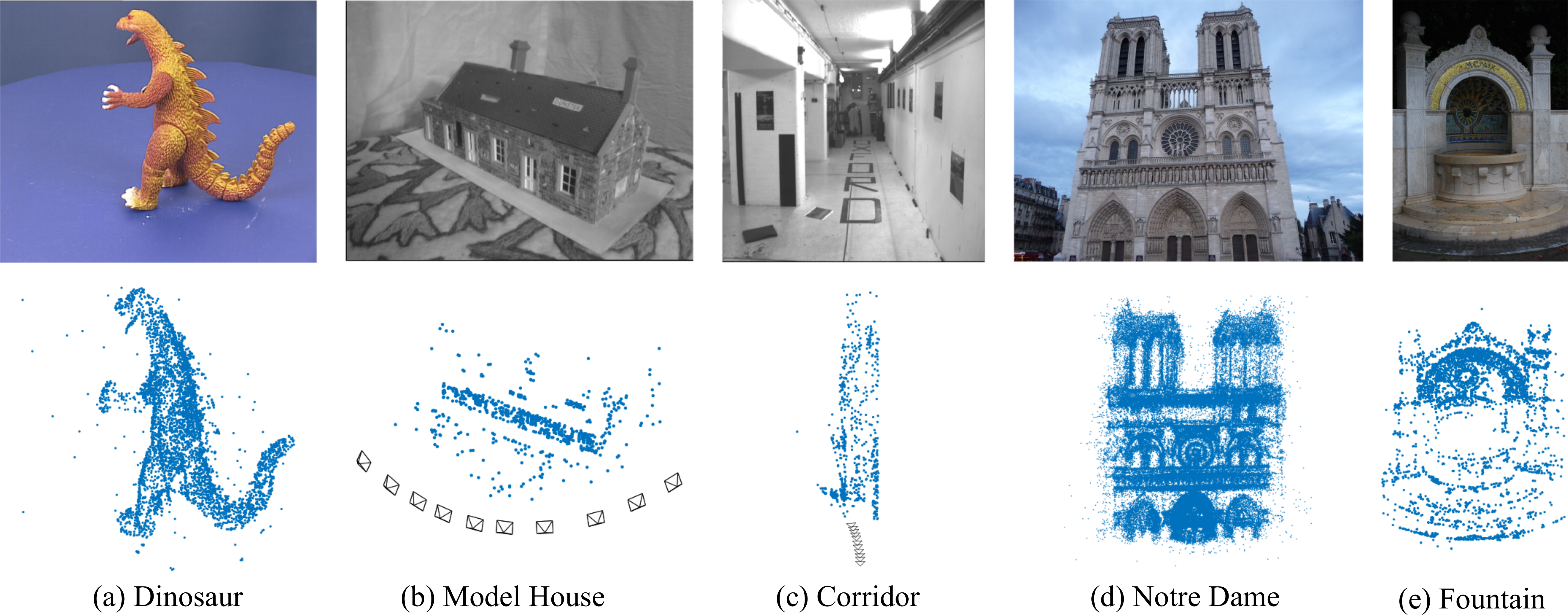}
\captionof{figure}{\textbf{Top row:} Real dataset images. \textbf{Bottom row:} Main segments of the median reconstruction results using the proposed $L_1$ method.
\label{fig:reconstruction}}
\end{figure*}

\begin{figure}[t]
 \centering
 \vspace{-1em}
 \includegraphics[width=0.45\textwidth]{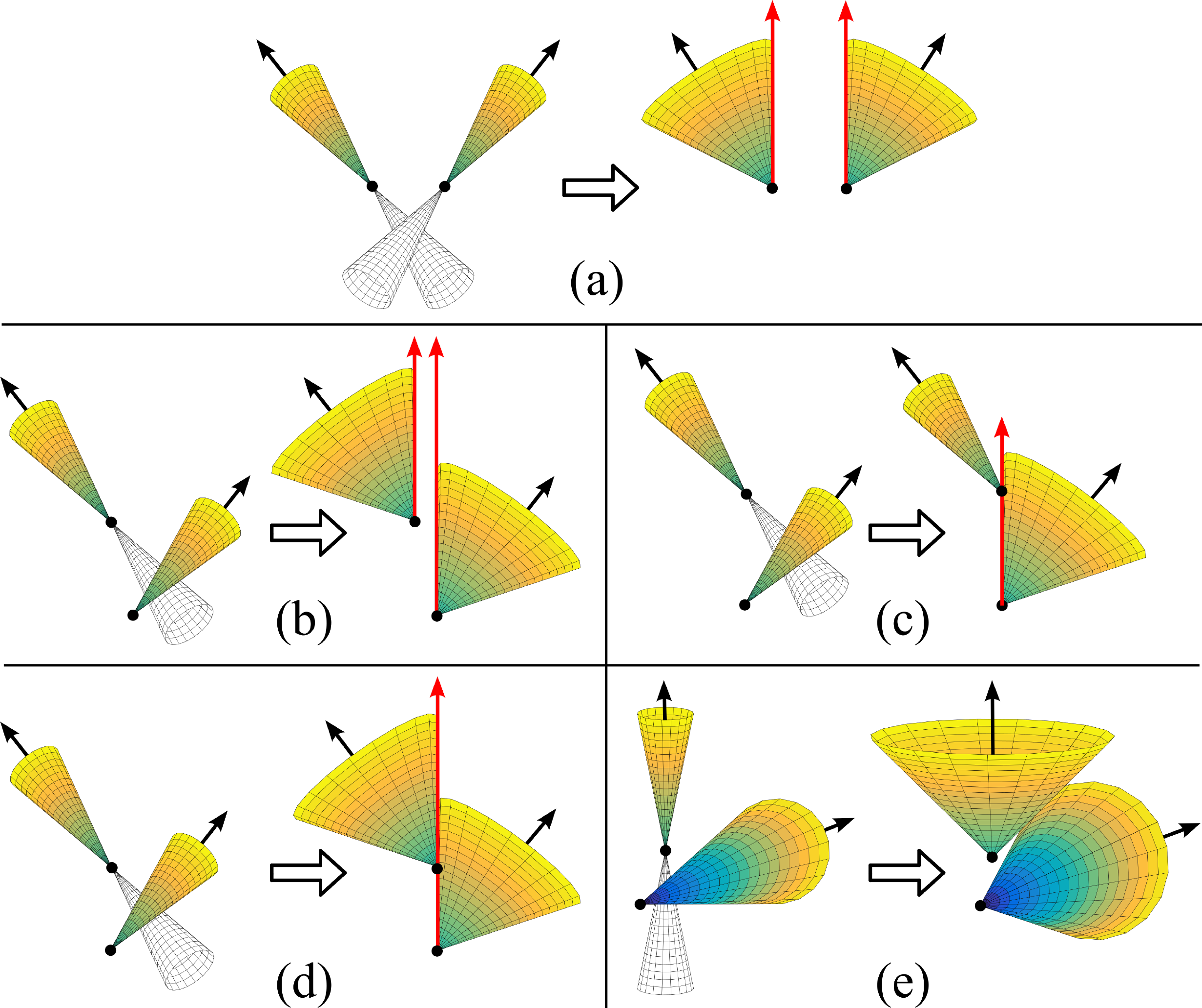}
\caption{Five scenarios where the optimal solution violates the cheirality constraint, and the possible reattempts for triangulation.
 }
\label{fig:cheirality}
\end{figure}

\section{Cheirality, Parallax and Outliers}
\label{sec:cheirality}
We have used the term \textit{lines} instead of \textit{rays} in all lemmas so far, ignoring the cheirality constraint \cite{hartley_book}.
We argue that if the optimal solution violates the cheirality constraint, the most reasonable choice is to simply discard the result.
In the following, we provide the rationale for this choice.

Fig. \ref{fig:cheirality} illustrates five scenarios where the optimal solution violates the cheirality constraint.
In case (a), both rays have negative depths at the optimal intersection.
Increasing the allowed angular reprojection error, the first intersection with positive depths occurs when the two corrected rays become parallel, resulting in a point at infinity.
This point cannot be triangulated, so it should be discarded.

In the remaining cases, the optimal intersection involves only one of the rays having a negative depth. 
Following the same procedure, the first intersection with positive depths occurs either at infinity (case (b)), at one of the camera centers (case (c)), along the ray parallel to the translation (case (d)), or at a point somewhere else (case (e)).

In case (b), (c) and (d), the newly triangulated point has either infinite, zero or ambiguous depth, so it is reasonable to discard it.
In case (e), we found that reattempting the triangulation with positive depths yields either a very large error, a point near the epipole or a low parallax angle.
Typically, these are the indicators of low accuracy or an outlier \cite{hartley_triangulation, hartley_book}, so a reasonable choice is to discard the match.
This procedure is outlined in Step 4--6 of Tab. \ref{tab:summary}.
\vspace{-0.2em}

\section{Experimental Results}
\label{sec:results}
We evaluate the proposed methods in comparison to the midpoint method \cite{midpoint, hartley_triangulation}, Hartley and Sturm's $L_1$ and $L_2$ method \cite{hartley_triangulation}, Lindstrom's $L_2$ method with five iterations \cite{lindstrom}, and N\'{i}ster's $L_\infty$ method \cite{nister_phd_thesis}.
The evaluation was performed on both synthetic and real datasets.
We generated the synthetic datasets as follows:
A set of $8\times4$ point clouds of 2,500 points each are generated with a Gaussian radial distribution $\mathcal{N}(0,(d/4)^2)$ where $d$ is the distance from the world origin.
Each point cloud is centered at $[0,0,d]^\intercal$ for $d=2^n$ with $n=-1,0,...,+6$, and their image projections are perturbed by Gaussian noise $\mathcal{N}(0,\sigma^2)$ for $\sigma=0.5,1,2,4,8$.
The size and the focal length of the images are $1,024 \times 1,024$ pixels and $512$ pixel, respectively.
We have three configurations for the camera poses: (1) ``orbital" - the cameras at $[\pm0.5,0,0]^\intercal$ pointing at the point cloud center, (2) ``lateral" - the cameras at $[\pm0.5,0,0]^\intercal$ pointing at $[0,0,\infty]^\intercal$, and (3) ``forward" - the cameras at $[0,0,\pm0.5]^\intercal$ pointing at the point cloud center.
The poses are slightly perturbed with uniform noise $\mathcal{U}(0, 0.01)$.
For real datasets, we used the Oxford Dinosaur, Model House and Corridor \cite{oxford_dataset}, Notre Dame \cite{notredame_dataset} and Fountain \cite{olsson_dataset1,olsson_dataset2} dataset.
In total, the synthetic and real datasets provide over 5.5 million unique triangulation problems in a wide variety of geometric configurations.

\begin{figure}[t]
 \centering
 \vspace{-1em}
 \includegraphics[width=0.475\textwidth]{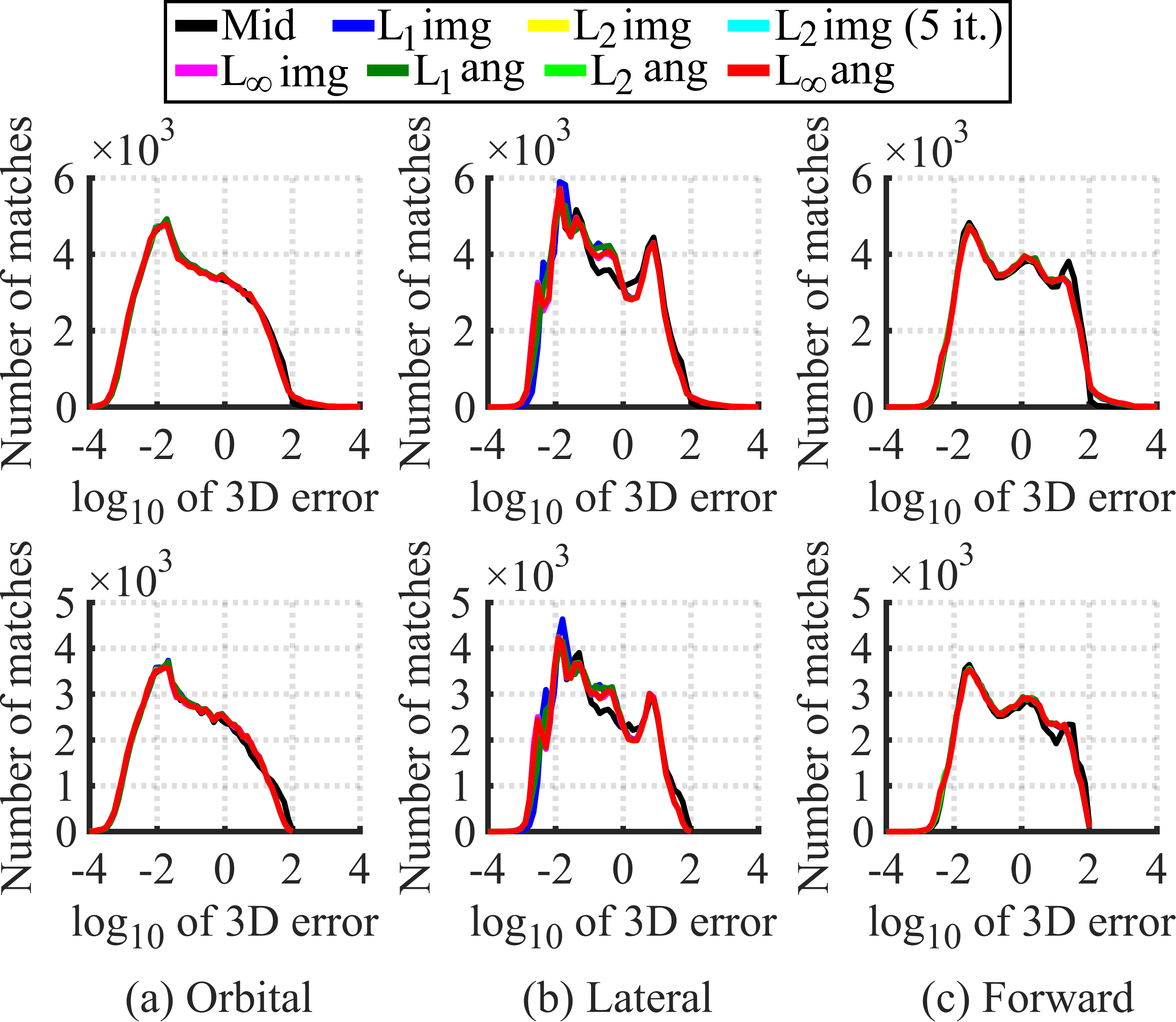}
\caption{3D triangulation errors before (top) and after (bottom) discarding the points with the lowest 5\% parallax.
 }
\label{fig:synthetic_results}
\end{figure}

{\renewcommand{\arraystretch}{1.2}%
\begin{table*}[t]
\small
\begin{center}
\begin{tabular}{cl|cccccccc|}
\cline{3-10}
&& Midpoint & $L_1$ img & $L_2$ img& $L_2$ img & $L_\infty$ img & \multirow{2}{*}{$L_1$ ang} & \multirow{2}{*}{$L_2$ ang} & \multirow{2}{*}{$L_\infty$ ang}\\
 &  &\cite{midpoint, hartley_triangulation}&\cite{hartley_triangulation} & \cite{hartley_triangulation}& 5 it. \cite{lindstrom} & \cite{nister_phd_thesis} &&&\\
\hline
\multicolumn{1}{|c|}{\parbox[t]{2mm}{\multirow{7}{*}{\rotatebox[origin=c]{90}{Error Criterion}}}} & $\theta_0+\theta_1$  & - & -  & -  & -  & - & \textbf{100 \%} & -  & -  \\
\multicolumn{1}{|c|}{}&$\theta_0^2+\theta_1^2$ & -  & -  & 7e-5 \%  & 5e-5 \% & - & - & \textbf{99.9999 \%}  & -  \\
\multicolumn{1}{|c|}{}&$\sin^2(\theta_0)+\sin^2(\theta_1)$ & -  & -  & -  & -  & - & - & \textbf{100 \%}  & -  \\
\multicolumn{1}{|c|}{}&$\max(\theta_0, \theta_1)$  & -  & -  & -  & -  & - & - & -  & \textbf{100 \%}  \\
\cline{2-10}
\multicolumn{1}{|c|}{}&$d_0+d_1$  & - & \textbf{70.84\%}& 0.002\%& 0.002\%& - & 29.16 \% & - & -\\
\multicolumn{1}{|c|}{}& $d_0^2+d_1^2$  & - & - & 23.14 \% & \textbf{76.86 \%}& -&- &- &- \\
\multicolumn{1}{|c|}{}& $\max(d_0, d_1)$  & -  & -  & -  & -  & \textbf{100 \%} & - & -  & -  \\ 
\hline
\end{tabular}
\end{center}
\caption{Percentage of the total matches (from all synthetic and real datasets) for which each method yields the lowest error in given criterion.
``img/ang": optimal in the image/angular errors.
See the supplementary material for the results from individual datasets.}
\label{tab:error_results}
\end{table*}
}
{\renewcommand{\arraystretch}{1.2}%
\begin{table*}[ht]
\small
\begin{center}
\begin{tabular}{cccccccccc|}
\cline{2-10}
 & \multicolumn{1}{|c}{Midpoint} & $L_1$ img & $L_2$ img  & $L_\infty$ img & $L_2$ img & $L_2$ img& \multirow{ 2}{*}{$L_1$ ang} & \multirow{ 2}{*}{$L_2$ ang} & \multirow{ 2}{*}{$L_\infty$ ang} \\
 &  \multicolumn{1}{|c}{\cite{midpoint, hartley_triangulation}}& \cite{hartley_triangulation} & \cite{hartley_triangulation} & \cite{nister_phd_thesis} & 2 it. \cite{lindstrom} & 5 it. \cite{lindstrom} & & & \\
\hline
\multicolumn{1}{|c|}{Points/sec}& 42 M & 65 K & 92 K& 270 K & 1.4 M& 520 K& 29 M& 670 K& 14 M\\
\multicolumn{1}{|c|}{Relative Speed} & 1.0 & 0.0016 & 0.0022 & 0.0064 &0.033 & 0.013 & 0.71 & 0.016 & 0.33 \\
\hline
\end{tabular}
\end{center}
\caption{Speed of computing a 3D point. The relative speed is normalized by that of the midpoint method. Note that this does not take into account Step 4--6 of Tab. \ref{tab:summary}. All algorithms were implemented in C++ and run on a laptop CPU (Intel i7-4810MQ, 2.8 GHz).}
\label{tab:timings}
\end{table*}
}

Tab. \ref{tab:error_results} provides the percentage of the total matches (from both synthetic and real datasets) for which each method yields the lowest error in given criterion.
In 100 \% of the total triangulation problems, all three of our methods yield the lowest errors in their corresponding optimal criterion.
We also see that minimizing $\sin^2(\theta_0)+\sin^2(\theta_1)$ is very close to minimizing $\theta_0^2+\theta_1^2$, as discussed in Section \ref{sec:L_2}.
Since our $L_1$ angular method is numerically stable, it sometimes finds better solutions than Hartley-Sturm's closed-form $L_1$ method \cite{hartley_triangulation} even in the $L_1$ image error criterion ($d_0+d_1$).

In Fig. \ref{fig:synthetic_results}, histograms are given for the 3D reconstruction errors on the synthetic datasets.
It shows that 1) all methods exhibit similar 3D accuracy, and 2) discarding low-parallax points (Step 6 of Tab. \ref{tab:summary}) helps to remove large 3D errors.
Qualitatively, we also found that the reconstructions of the real datasets look similar for all methods.
Fig. \ref{fig:reconstruction} shows the reconstruction results using the proposed $L_1$ method.

We compare the speed of each algorithm in Tab. \ref{tab:timings}.  
The midpoint method is the fastest, as it directly computes the 3D point using \eqref{eq:3d_point} without correcting the feature rays or image points. 
Among the optimal methods, our $L_1$ and $L_\infty$ methods are significantly faster than the rest, i.e., at least 1--2 orders of magnitude faster than the state-of-the-art \cite{lindstrom}.

\vfill

\section{Conclusions}
\label{sec:conclusion}
In this work, we derived optimal closed-form solutions to the $L_1$, $L_2$ and $L_\infty$ stereo triangulation based on the angular reprojection error.
The proposed triangulation methods are extremely simple and fast, and they guarantee global optimality under respective cost functions.
We believe that our findings will be particularly useful for large-scale SfM and real-time visual SLAM algorithms.

\section*{Acknowledgement}
This work was partially supported by the Spanish government (project PGC2018-096367-B-I00) and the Arag{\'{o}}n regional government (Grupo DGA-T45{\_}17R/FSE).

\renewcommand{\thesubsection}{\Alph{subsection}}
\section*{Appendix}
\subsection{Proof of Lemma \ref{lemma:single_pivot}}
\label{app:proof_single_pivot}
\begin{figure}[t]
 \centering
 \includegraphics[width=0.25\textwidth]{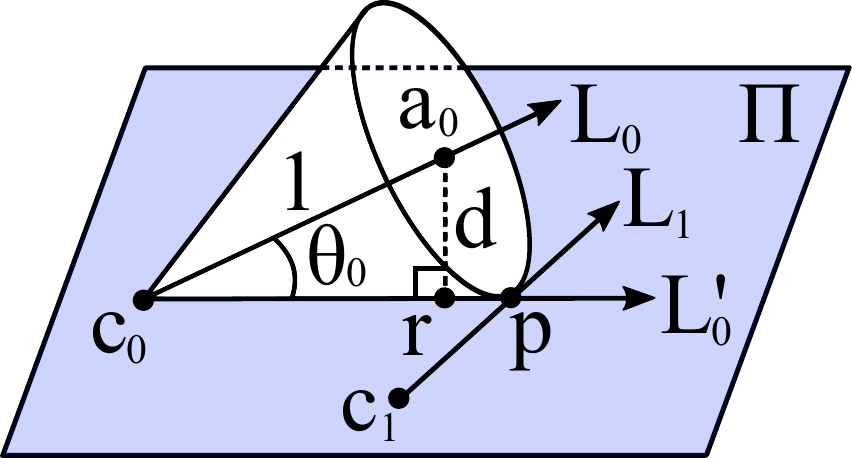}
\caption{The angle $\theta_0$ is the smallest angle required for pivoting line $\mathbf{L}_0$ at point $\mathbf{c}_0$ to make it intersect $\mathbf{L}_1$. 
}
\label{fig:cone_plane_min_single}
\end{figure}
Consider a right circular cone with apex $\mathbf{c}_0$ and axis $\mathbf{L}_0$, lying sideways on a plane $\Pi$ that contains $\mathbf{c}_0$ and line $\mathbf{L}_1$ (see Fig. \ref{fig:cone_plane_min_single}).
The equation of the plane is given by 
\vspace{-0.2em}
\begin{equation}
    \Pi(\mathbf{x})=
    \mathbf{n}_1\cdot(\mathbf{x}-\mathbf{c}_0)=0
    \quad\text{with}\quad
    \mathbf{n}_1 = \mathbf{m}_1\times\mathbf{t}.
    \vspace{-0.2em}
\end{equation}
The line of intersection between the plane and the cone forms the smallest angle to $\mathbf{L}_0$ among all possible lines on the plane that pass $\mathbf{c}_0$.
That is, it forms the smallest angle to $\mathbf{L}_0$ among all possible lines that pass both $\mathbf{c}_0$ and $\mathbf{L}_1$.
Hence, this line of intersection must be $\mathbf{L}'_0$.
Now, consider a point $\mathbf{a}_0 = \mathbf{c}_0+\widehat{\mathbf{m}}_0$ located one unit away from $\mathbf{c}_0$ along $\mathbf{L}_0$.
Let $\mathbf{r}$ be the projection of $\mathbf{a}_0$ onto plane $\Pi$.
According to lemma \ref{lemma:cone-on-plane-perpendicularity} in Appendix \ref{app:proof_others}, the point $\mathbf{r}$ must be located along $\mathbf{L}'_0$.
Let $d=\lVert\mathbf{a}_0-\mathbf{r}\rVert$, i.e., the distance between $\mathbf{a}_0$ and plane $\Pi$.
Then, we obtain $\sin{(\theta_0)}$ as follows:
\vspace{-0.5em}
\begin{equation*}
    \sin{(\theta_0)} 
    = d
    \stackrel{\eqref{eq:shortest_distance_point_plane}}{=}
    |\widehat{\mathbf{n}}_1\cdot(\mathbf{a}_0-\mathbf{c}_0)|
    =
    |\widehat{\mathbf{n}}_1\cdot\widehat{\mathbf{m}}_0|.
    \tag*{$\blacksquare$}
    \vspace{-0.5em}
\end{equation*}\vspace{-1em}

\subsection{Proof of Lemma \ref{lemma:L_1}}
\label{app:proof_L_1}
One of the following is true when $(\theta_0+\theta_1)$ is minimized:
\begin{enumerate}\itemsep-2pt
    \item \label{scenario1} $\mathbf{L}_0'\neq \mathbf{L}_0$ and $\mathbf{L}_1'=\mathbf{L}_1 \longleftrightarrow \theta_0>0$ and $\theta_1=0$.
    \item \label{scenario2} $\mathbf{L}_0'=\mathbf{L}_0$ and $\mathbf{L}_1'\neq \mathbf{L}_1 \longleftrightarrow \theta_0=0$ and $\theta_1>0$.
    \item \label{scenario3}$\mathbf{L}_0'\neq \mathbf{L}_0$ and $\mathbf{L}_1'\neq\mathbf{L}_1 \longleftrightarrow \theta_0>0$ and $\theta_1>0$.
\end{enumerate}\vspace{-0.5em}

Suppose, for the sake of argument, that one of the first two statements is true. 
In the first case, lemma \ref{lemma:single_pivot} states that $\mathbf{m}'_0$ is obtained by projecting $\mathbf{m}_0$ onto the plane with the normal $\mathbf{m}_1\times\mathbf{t}$, which leads to \eqref{eq:L_1:1} and
\vspace{-0.2em}
\begin{equation}
\label{eq:proof_L_1:1}
    \sin(\theta_0) 
    =
    \frac{|\widehat{\mathbf{m}}_0\cdot(\mathbf{m}_1\times\mathbf{t})|}{\lVert\mathbf{m}_1\times\mathbf{t}\rVert}
    =
    \frac{|\widehat{\mathbf{m}}_0\cdot(\widehat{\mathbf{m}}_1\times\mathbf{t})|}{\lVert\widehat{\mathbf{m}}_1\times\mathbf{t}\rVert}.
    \vspace{-0.2em}
\end{equation}
Likewise, in the second case, lemma \ref{lemma:single_pivot} leads to \eqref{eq:L_1:2}, and 
\vspace{-0.2em}
\begin{equation}
\label{eq:proof_L_1:2}
    \sin(\theta_1) 
    =
    \frac{|\widehat{\mathbf{m}}_1\cdot(\mathbf{m}_0\times\mathbf{t})|}{\lVert\mathbf{m}_0\times\mathbf{t}\rVert}
    \stackrel{\eqref{eq:triple_product_invariance}}{=}
    \frac{|\widehat{\mathbf{m}}_0\cdot(\widehat{\mathbf{m}}_1\times\mathbf{t})|}{\lVert\widehat{\mathbf{m}}_0\times\mathbf{t}\rVert}.
    \vspace{-0.2em}
\end{equation}
Now, the question is how to determine which of the two statements is true.
Comparing the right-hand side of \eqref{eq:proof_L_1:1} and \eqref{eq:proof_L_1:2}, we find that if $\lVert\widehat{\mathbf{m}}_0\times\mathbf{t}\rVert \leq \lVert\widehat{\mathbf{m}}_1\times\mathbf{t}\rVert$, then
\vspace{-0.2em}
\begin{equation}
\label{eq:proof_L_1:3}
    \min_{\scaleto{\theta_0| \theta_1=0}{8pt}} \ \theta_0 \leq \min_{\scaleto{\theta_1| \theta_0=0}{8pt}} \ \theta_1,
    \vspace{-0.2em}
\end{equation}
and $\min(\theta_0+\theta_1)$ is equal to the left-hand side of \eqref{eq:proof_L_1:3}, indicating that the first statement is true.
Naturally, the second statement is true otherwise.
Note that there is an ambiguity if $\lVert\widehat{\mathbf{m}}_0\times\mathbf{t}\rVert = \lVert\widehat{\mathbf{m}}_1\times\mathbf{t}\rVert$, and the solution is optimal whichever case is considered.
This concludes the proof of lemma \ref{lemma:L_1} for the first two cases.

\begin{figure}[t]
 \centering
 \includegraphics[width=0.36\textwidth]{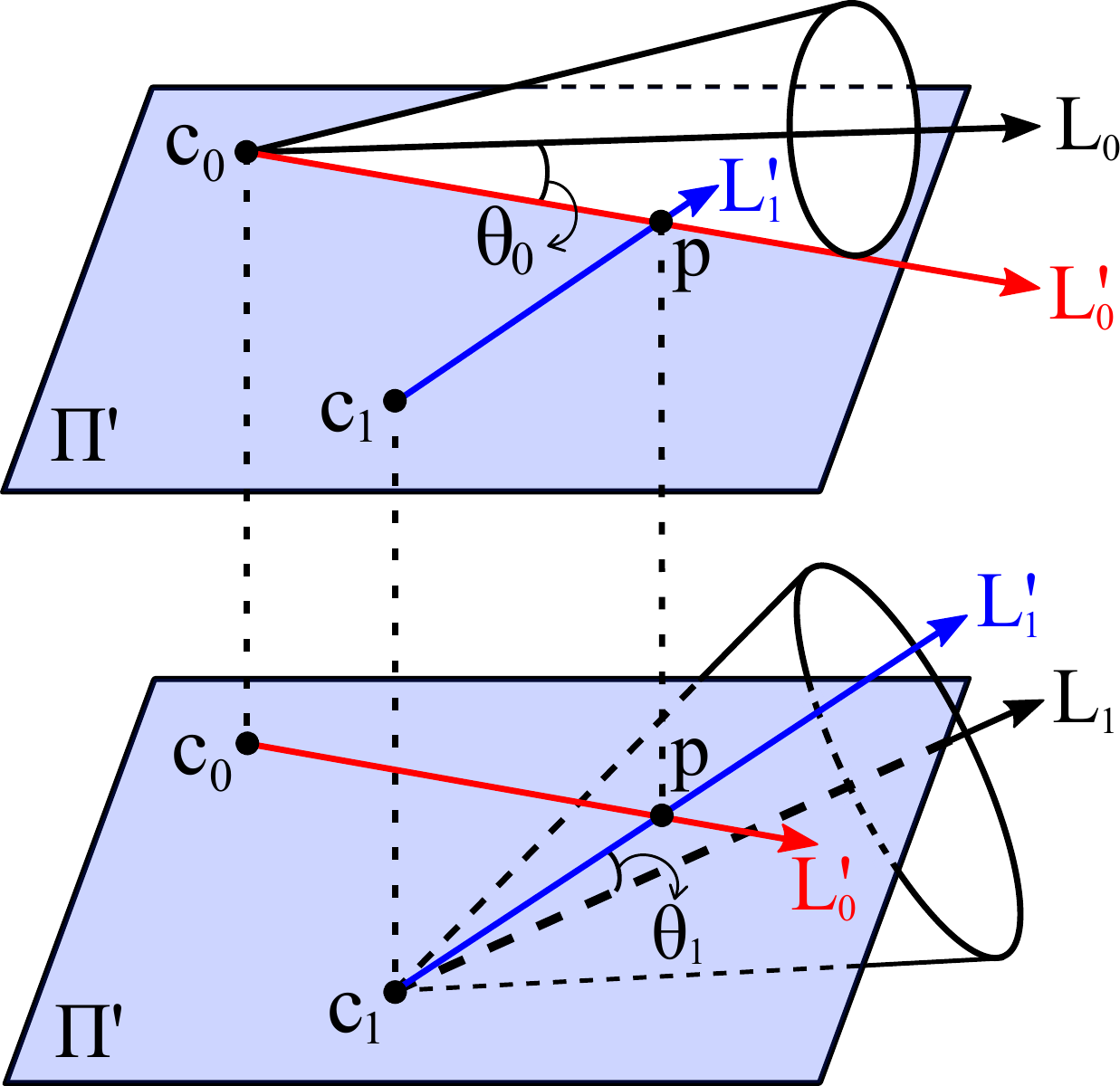}
\caption{When two cones intersect at a single point $\mathbf{p}$ on their lateral surface, they are tangent to the same plane containing $\mathbf{p}$ and the apexes of each cone.
For visualization purposes, we show the two cones lying on each side of the plane separately.
 }
\label{fig:cones_plane}
\end{figure}

We will now prove that the third case never occurs.
Given some angle $\theta_1$, minimizing $(\theta_0+\theta_1)$ is equivalent to minimizing $\theta_0$.
This is the identical situation as the first case if we replace $\mathbf{L}_1$ by $\mathbf{L}_1'$.
We know from the proof of lemma \ref{lemma:single_pivot} that pivoting a line to intersect another with minimum angle can be modeled by a cone lying sideways on a plane.
The top of Fig. \ref{fig:cones_plane} illustrates this.
Similarly, the bottom of Fig. \ref{fig:cones_plane} illustrates the minimization of $\theta_1$ with respect to $\mathbf{L}_0'$ given $\theta_0$.
Now, since both planes touching each cone contain the same two intersecting lines $\mathbf{L}_0'$ and $\mathbf{L}_1'$, they must be the same plane.
Let this plane be $\Pi'$.
According to lemma \ref{lemma:single_pivot}, $\mathbf{L}_0'$ is the projection of $\mathbf{L}_0$ onto plane $\Pi'$.
Therefore, 
\vspace{-1.3em}
\begin{equation}
\label{eq:lemma_min_pivot_angle_proof3}
    \widehat{\mathbf{m}}_0' = \frac{\widehat{\mathbf{m}}_0-(\widehat{\mathbf{m}}_0\cdot\widehat{\mathbf{n}}')\widehat{\mathbf{n}}'}{\lVert\widehat{\mathbf{m}}_0-(\widehat{\mathbf{m}}_0\cdot\widehat{\mathbf{n}}')\widehat{\mathbf{n}}'\rVert}, 
\end{equation}
where $\widehat{\mathbf{n}}'$ is the unit normal of plane $\Pi'$.
Since $\Pi'$ contains both $\mathbf{c}_0$ and $\mathbf{c}_1$, $\widehat{\mathbf{n}}'$ is perpendicular to $\mathbf{t}=\mathbf{c}_0-\mathbf{c}_1$.
Hence, computing the dot product with $\mathbf{t}$ on each side of \eqref{eq:lemma_min_pivot_angle_proof3} yields
\begin{equation}
\label{eq:lemma_min_pivot_angle_proof4}
\mathbf{t}\cdot\widehat{\mathbf{m}}_0' = \frac{\mathbf{t}\cdot\widehat{\mathbf{m}}_0}{\lVert\widehat{\mathbf{m}}_0-(\widehat{\mathbf{m}}_0\cdot\widehat{\mathbf{n}}')\widehat{\mathbf{n}}'\rVert}.
\end{equation}
Note that $\lVert\widehat{\mathbf{m}}_0-(\widehat{\mathbf{m}}_0\cdot\widehat{\mathbf{n}}')\widehat{\mathbf{n}}'\rVert$ corresponds to the magnitude of the projection of $\widehat{\mathbf{m}}_0$ onto plane $\Pi'$ for non-zero $\theta_0$, so it must be smaller than $\lVert\widehat{\mathbf{m}}_0\rVert =1$.
Thus,
\begin{equation}
\label{eq:lemma_min_pivot_angle_proof5}
|\mathbf{t}\cdot\widehat{\mathbf{m}}_0'| = \frac{|\mathbf{t}\cdot\widehat{\mathbf{m}}_0|}{\lVert\widehat{\mathbf{m}}_0-(\widehat{\mathbf{m}}_0\cdot\widehat{\mathbf{n}}')\widehat{\mathbf{n}}'\rVert} > |\mathbf{t}\cdot\widehat{\mathbf{m}}_0|.
\end{equation}
Using \eqref{eq:|axb|^2}, this inequality can be written as
\vspace{-0.5em}
\begin{equation}
\label{eq:lemma_min_pivot_angle_proof6}
\lVert\mathbf{t}\times\widehat{\mathbf{m}}_0'\rVert < \lVert\mathbf{t}\times\widehat{\mathbf{m}}_0\rVert.
\vspace{-0.5em}
\end{equation}
Analogously, we can also derive 
\vspace{-0.5em}
\begin{equation}
\label{eq:lemma_min_pivot_angle_proof7}
\lVert\mathbf{t}\times\widehat{\mathbf{m}}_1'\rVert < \lVert\mathbf{t}\times\widehat{\mathbf{m}}_1\rVert.
\vspace{-0.5em}
\end{equation}
Now, suppose that 
\vspace{-0.5em}
\begin{equation}
\label{eq:lemma_min_pivot_angle_proof7.1}
    \min_{\scaleto{\theta_0,\theta_1}{8pt}} \ (\theta_0+\theta_1) = \theta_0^*+\theta_1^* \quad \text{with} \quad \theta_0^*, \ \theta_1^*>0.
    \vspace{-0.5em}
\end{equation}
Without loss of generality, let us assume that $\lVert\mathbf{t}\times\widehat{\mathbf{m}}_0\rVert \leq \lVert\mathbf{t}\times\widehat{\mathbf{m}}_1\rVert$.
Then, \eqref{eq:lemma_min_pivot_angle_proof6} gives $\lVert\mathbf{t}\times\widehat{\mathbf{m}}_0'\rVert < \lVert\mathbf{t}\times\widehat{\mathbf{m}}_1\rVert$.
As we discussed for the first two cases, this means that pivoting $\mathbf{L}_0'$ to intersect $\mathbf{L}_1$ takes smaller angle than pivoting $\mathbf{L}_1$ to intersect $\mathbf{L}_0'$, i.e., $\theta_0' < \theta_1^*$. Thus
\vspace{-0.3em}
\begin{equation}
\label{eq:lemma_min_pivot_angle_proof8}
    \theta_0^*+\theta_0' < \theta_0^*+\theta_1^*.
    \vspace{-0.3em}
\end{equation}
According to lemma \ref{lemma:single_vs_multi} in Appendix \ref{app:proof_others}, pivoting a line twice for intersection takes equal or greater angle than the single minimum pivot angle.
Therefore, 
\vspace{-0.5em}
\begin{equation}
\label{eq:lemma_min_pivot_angle_proof9}
    \min_{\scaleto{\theta_0|\theta_1=0}{8pt}} \ \theta_0 \leq \theta_0^*+\theta_0' < \theta_0^*+\theta_1^*,
    \vspace{-0.5em}
\end{equation}
which contradicts \eqref{eq:lemma_min_pivot_angle_proof7.1}.
Therefore, $(\theta_0+\theta_1)$ is minimized when either $\theta_0$ or $\theta_1$ is zero. \QEDA

\subsection{Proof of Lemma \ref{lemma:L_2}}
\label{app:proof_L_2}
Given some angle $\theta_1$, $\left(\sin^2{(\theta_0)}+\sin^2{(\theta_1)}\right)_\text{min}$ is achieved by minimizing $\theta_0$ and vice versa.
As discussed in the proof of lemma \ref{lemma:L_1}, this means that the underlying geometry at $\left(\sin^2{(\theta_0)}+\sin^2{(\theta_1)}\right)_\text{min}$ can be represented by the two cones with apex $\mathbf{c}_0$, $\mathbf{c}_1$ and skew axes $\mathbf{L}_0$, $\mathbf{L}_1$, respectively, touching each side of the same plane on their lateral surface.
This is visualized in Fig. \ref{fig:cones_plane}.
Let ${\mathbf{n}}'$ be the normal of plane $\Pi'$.
From Lemma \ref{lemma:single_pivot}, we know that
\vspace{-0.5em}
\begin{equation}
    \sin{(\theta_0)} 
    =
    |\widehat{\mathbf{n}}'\cdot\widehat{\mathbf{m}}_0|
    \quad\text{and}\quad
    \sin(\theta_1)
    =|\widehat{\mathbf{n}}'\cdot\widehat{\mathbf{m}}_1|.
    \vspace{-0.5em}
\end{equation}
Combining these two equations, we get
\vspace{-0.5em}
\begin{equation}
    \begin{gathered}
            \sin^2{(\theta_0)}+\sin^2{(\theta_1)} 
    =
    \lVert\mathbf{M}\strut^\intercal\widehat{\mathbf{n}}'\rVert^2 \\
    \text{with} \quad \mathbf{M} = \begin{bmatrix}\widehat{\mathbf{m}}_0 & \widehat{\mathbf{m}}_1\end{bmatrix}.
    \end{gathered}
    \vspace{-0.5em}
\end{equation}
Since plane $\Pi'$ contains both $\mathbf{c}_0$ and $\mathbf{c}_1$, $\widehat{\mathbf{n}}'$ is perpendicular to $\mathbf{t}=\mathbf{c}_0-\mathbf{c}_1$.
Therefore, minimizing $(\sin^2{(\theta_0)}+\sin^2{(\theta_1)})$ is equivalent to solving the following equality-constrained quadratic programming problem:
\vspace{-0.5em}
\begin{equation}
    \label{eq:lemma:L_2:proof1}
    \argmin_{\widehat{\mathbf{n}}'} \ \lVert\mathbf{M}\strut^\intercal\widehat{\mathbf{n}}'\rVert^2, \ \text{s.t.} \ \lVert\widehat{\mathbf{n}}'\rVert=1 \ \text{and} \ \mathbf{t}\cdot\widehat{\mathbf{n}}'=0.
    \vspace{-0.5em}
\end{equation}
In \cite{golub_eigenvalue}, it was shown that this problem can be solved using the method of Lagrange multipliers, and $\lVert\mathbf{M}\strut^\intercal\widehat{\mathbf{n}}'\rVert^2$ is minimized when $\widehat{\mathbf{n}}'$ is the eigenvector corresponding to the smallest nontrivial eigenvalue of $\mathbf{A} = (\mathbf{I}-\widehat{\mathbf{t}}\ \widehat{\mathbf{t}}\strut^\intercal)\mathbf{MM}\strut^\intercal$.
Letting $\mathbf{P}=(\mathbf{I}-\widehat{\mathbf{t}}\ \widehat{\mathbf{t}}\strut^\intercal)$, it can be easily shown that 
$\mathbf{P} = \mathbf{P}\strut^\intercal = \mathbf{P}\strut^\intercal\mathbf{P}=\mathbf{P}\mathbf{P}\strut^\intercal$.
Hence,  $\mathbf{A}=\mathbf{PMM}\strut^\intercal=\mathbf{P}\mathbf{P}\strut^\intercal\mathbf{MM}\strut^\intercal$.
Note that for any square matrix $\mathbf{X}$ and $\mathbf{Y}$, the eigen-decomposition of $\mathbf{XY}$ is the same as that of $\mathbf{YX}$.
This means that the eigenvectors of $\mathbf{A}=\mathbf{P}\left(\mathbf{P}\strut^\intercal\mathbf{MM}\strut^\intercal\right)$ are the same as those of $\left(\mathbf{P}\strut^\intercal\mathbf{MM}\strut^\intercal\right)\mathbf{P}=(\mathbf{M}\strut^\intercal\mathbf{P})\strut^\intercal(\mathbf{M}\strut^\intercal\mathbf{P})$, i.e., the right-singular vectors of $\mathbf{M}\strut^\intercal\mathbf{P}$.
Therefore, letting $\mathbf{USV}\strut^\intercal = \text{SVD}\left(\mathbf{M}\strut^\intercal\mathbf{P}\right)$ with the diagonal entries of $\mathbf{S}$ in descending order, the optimal $\widehat{\mathbf{n}}'$ is given by the second column of $\mathbf{V}$.
Finally, projecting $\mathbf{m}_0$ and $\mathbf{m}_1$ onto plane $\Pi'$ leads to \eqref{eq:lemma:L_2:1}. \QEDA

\subsection{Proof of Lemma \ref{lemma:L_inf}}
\label{app:proof_L_inf}
First, we show that $\theta_0=\theta_1$ when $\max(\theta_0, \theta_1)$ is minimized:
Consider two cones with apex $\mathbf{c}_0$, $\mathbf{c}_1$ and skew axes $\mathbf{L}_0$, $\mathbf{L}_1$. 
Constrain both their apertures to be $2\theta$.
When $\theta=0$, the they are simply two skew lines.
As we gradually increase $\theta$, they will grow at the same rate, and eventually, touch one another. 
Let $\theta=\theta'$ at this point.
Now, suppose
\vspace{-0.4em}
\begin{equation}
\label{eq:L_inf_proof1}
\theta^*
:=
\min_{\scaleto{\theta_0, \theta_1}{8pt}}\max(\theta_0, \theta_1) 
< \theta'.
\vspace{-0.4em}
\end{equation}
The definition of $\theta^*$ implies that setting $\theta_0=\theta_1 = \theta^*$ will make the two cones partially overlap in space (or at least meet at a point).
However, the they do not meet when $\theta_0=\theta_1<\theta'$.
This is a contradiction, so the inequality in \eqref{eq:L_inf_proof1} must be false, and $\theta^*$ must be equal to $\theta'$.
That is,  $\theta_0=\theta_1=\theta'$  in order for $\max(\theta_0, \theta_1)$ to be minimized. 

We can now represent the underlying geometry at $(\max(\theta_0, \theta_1))_\text{min}$ as two congruent cones with skew axes, touching each side of the same plane $\Pi'$ on their lateral surface.
This is the situation shown in Fig. \ref{fig:cones_plane} for $\theta_0=\theta_1$.
Let ${\mathbf{n}}'$ be the normal of plane $\Pi'$.
Then, from lemma \ref{lemma:single_pivot}, we get
\vspace{-0.3em}
\begin{equation}
\label{eq:lemma:L_inf:proof1}
    \sin{(\theta_0)} 
    =
    \sin{(\theta_1)}
    =
    |\widehat{\mathbf{n}}'\cdot\widehat{\mathbf{m}}_0|
    =|\widehat{\mathbf{n}}'\cdot\widehat{\mathbf{m}}_1|.
    \vspace{-0.3em}
\end{equation}
The last equality in \eqref{eq:lemma:L_inf:proof1} can be written as
\vspace{-0.5em}
\begin{equation}
\label{eq:lemma:L_inf:proof2}
    (\widehat{\mathbf{m}}_0+w\widehat{\mathbf{m}}_1)\cdot\widehat{\mathbf{n}}'=0,
    \vspace{-0.5em}
\end{equation}
where $w$ is $-1$ or $1$, depending on the signs of $\widehat{\mathbf{n}}'\cdot\widehat{\mathbf{m}}_0$ and $\widehat{\mathbf{n}}'\cdot\widehat{\mathbf{m}}_1$.
On the other hand, since plane $\Pi'$ contains both $\mathbf{c}_0$ and $\mathbf{c}_1$, $\widehat{\mathbf{n}}'$ is perpendicular to $\mathbf{t}=\mathbf{c}_0-\mathbf{c}_1$: 
\vspace{-0.5em}
\begin{equation}
\label{eq:lemma:L_inf:proof3}
    \mathbf{t}\cdot\widehat{\mathbf{n}}'=0.
    \vspace{-0.5em}
\end{equation}
Combining \eqref{eq:lemma:L_inf:proof2} and \eqref{eq:lemma:L_inf:proof3}, $\widehat{\mathbf{n}}'$ can be expressed as
\vspace{-0.5em}
\begin{equation}
\label{eq:lemma:L_inf:proof4}
    \widehat{\mathbf{n}}'= \lambda(\widehat{\mathbf{m}}_0+w\widehat{\mathbf{m}}_1)\times\mathbf{t}
    \vspace{-0.5em}
\end{equation}
where $\lambda$ is the normalizing factor.
Evaluating \eqref{eq:lemma:L_inf:proof4} at $w=1$ and $w=-1$ gives two candidates for optimal $\widehat{\mathbf{n}}'$.
The optimal solution is then determined by comparing the values of \eqref{eq:lemma:L_inf:proof1} with each candidate $\widehat{\mathbf{n}}'$, which amounts to choosing the solution with smaller $\lambda$.
This procedure corresponds to \eqref{eq:lemma:L_inf:2}.
Finally, projecting $\mathbf{m}_0$ and $\mathbf{m}_1$ onto plane $\Pi'$ with optimal $\widehat{\mathbf{n}}'$ leads to \eqref{eq:lemma:L_inf:1}.
\QEDA

\begin{figure}[t]
 \centering
 \includegraphics[width=0.475\textwidth]{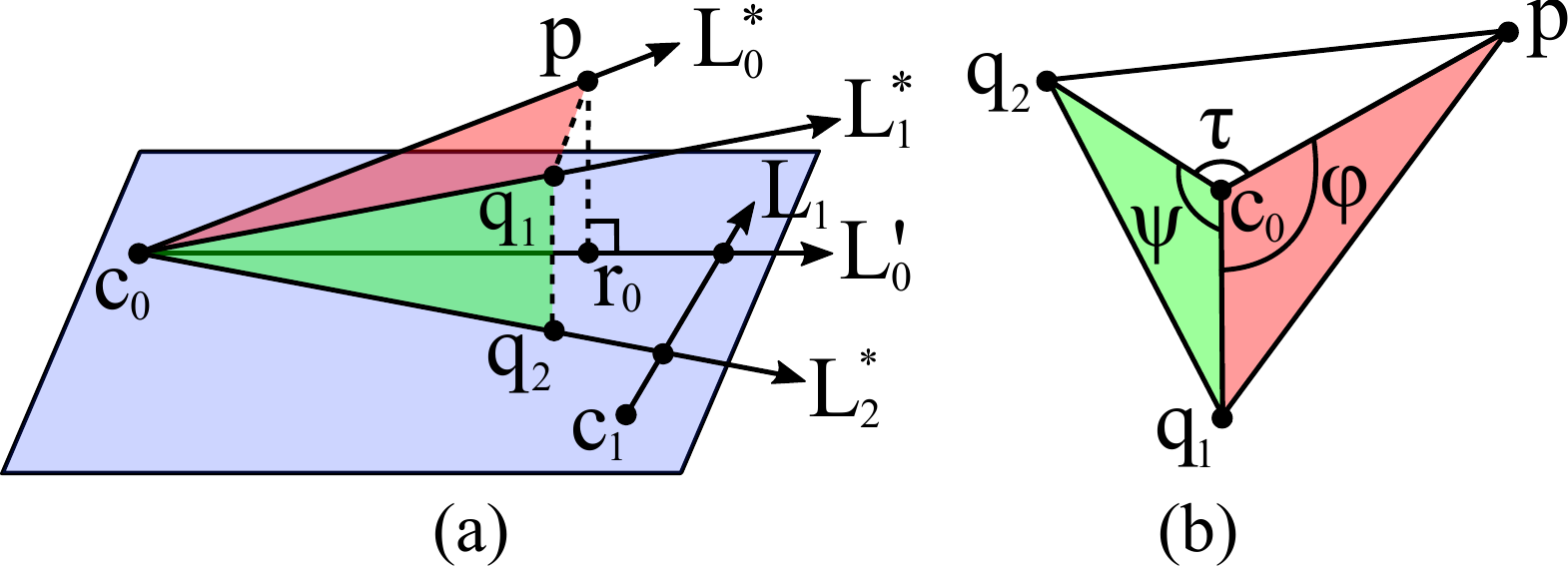}
\caption{\textbf{(a)} Pivoting line $\mathbf{L}_0^*$ in two steps ($\mathbf{L}_0^*\rightarrow\mathbf{L}_1^*\rightarrow\mathbf{L}_2^*$) to make it intersect to another line $\mathbf{L}_1$.
\textbf{(b)} A tetrahedron formed by point $\mathbf{c}_0$, $\mathbf{p}$, $\mathbf{q}_1$ and $\mathbf{q}_2$.
 }
\label{fig:multi_pivot}
\end{figure}

\subsection{Other Geometric Lemmas}
\label{app:proof_others}
\begin{lemma}[Cone-On-Plane Perpendicularity]
\label{lemma:cone-on-plane-perpendicularity}
When a plane is tangent to a right circular cone, the line of intersection is the projection of cone's axis onto the plane.
\end{lemma}
\vspace{-0.5em}
\noindent\textbf{Proof. }  
Consider a cone with axis $\mathbf{L}_0$ and plane $\mathbf{\Pi}$ tangent to this cone.
They are both symmetric with respect to the plane that contains $\mathbf{L}_0$ and the normal of $\mathbf{\Pi}$.
Let this plane of symmetry be $\mathbf{\Pi}_\text{sym}$.
For any circular cross-section of the cone, there is a single point touching $\mathbf{\Pi}$.
Therefore, this point must lie on $\mathbf{\Pi}_\text{sym}$, and so must the line of intersection, $\mathbf{L}_0'$.
It follows that $\mathbf{\Pi}_\text{sym}$ contains $\mathbf{L}_0$ and $\mathbf{L}_0'$.
Since $\mathbf{\Pi}_\text{sym}$ is perpendicular to $\mathbf{\Pi}$, $\mathbf{L}'_0$ is a projection of $\mathbf{L}_0$ onto $\mathbf{\Pi}$. \QEDA

\begin{lemma}[Single vs Multi-Pivot for Intersection]
\label{lemma:single_vs_multi}
Given two skew lines $\mathbf{L}_0(s_0) = \mathbf{c}_0+s_0{\mathbf{m}}_0$ and $\mathbf{L}_1(s_1) = \mathbf{c}_1+s_1{\mathbf{m}}_1$,
let $\mathbf{L}_0'$ be the line that forms the smallest angle $\theta_0\in[0, \ \pi/2]$ to $\mathbf{L}_0$ among all possible lines that intersect both point $\mathbf{c}_0$ and line $\mathbf{L}_1$.
For any positive integer $N$, consider the following arbitrary lines passing $\mathbf{c}_0$ such that
\vspace{-0.5em}
\begin{equation*}
    \mathbf{L}_{i}^*(s_i^*) = 
    \begin{cases}
    \mathbf{L}_0 \quad &\text{for} \quad i = 0 \\
    \mathbf{c}_0 + s_i^*{\mathbf{m}}_i^* \quad &\text{for} \quad i = 1,2, \cdots,N 
    \end{cases}
    \vspace{-0.5em}
\end{equation*}
where only $\mathbf{L}_{N}^*$ intersects $\mathbf{L}_1$.
Then,
\vspace{-0.5em}
\begin{equation}
\label{eq:lemma_single_vs_multi}
    \theta_0 \leq \sum_{i=1}^{N}\angle(\mathbf{L}_i^*, \ \mathbf{L}_{i-1}^*).
\end{equation}
\end{lemma}
\vspace{-0.5em}
\noindent\textbf{Proof. }  
The right-hand side of \eqref{eq:lemma_single_vs_multi} corresponds to the sum of $N$ pivot angles that make line $\mathbf{L}_0$ to intersect $\mathbf{L}_1$.
Fig. \ref{fig:multi_pivot}{\color{red}a} depicts such operation for $N=2$.
Let $\phi=\angle\left(\mathbf{L}_0^*, \ \mathbf{L}_1^*\right)$, $\psi=\angle\left(\mathbf{L}_1^*, \ \mathbf{L}_2^*\right)$ and $\tau=\angle\left(\mathbf{L}_0^*, \ \mathbf{L}_2^*\right)$. 
Now, consider three arbitrary points $\mathbf{p}$, $\mathbf{q}_1$ and $\mathbf{q}_2$ on $\mathbf{L}_0^*$, $\mathbf{L}_1^*$ and $\mathbf{L}_2^*$, respectively. 
A tetrahedron formed by these three points and $\mathbf{c}_0$ are shown in Fig. \ref{fig:multi_pivot}{\color{red}b}.
At a vertex of a tetrahedron, the three edges form three angles such that the sum of any two angles is greater than the third one \cite{solid_geometry, tetrahedron1}. 
Thus, $\tau \leq \phi + \psi$.
Since $\theta_0$ is the minimum pivot angle for intersection, we have $\theta_0\leq\tau\leq\phi+\psi$,
which proves \eqref{eq:lemma_single_vs_multi} for $N=2$. 
Now, for $N>2$, we know that replacing the last two pivots by the corresponding single minimum pivot will produce $N-1$ pivots that take equal or smaller angle.
Repeating this process until $N=1$ proves \eqref{eq:lemma_single_vs_multi} for any $N>2$. \QEDA

\balance
{\small
\bibliographystyle{ieee}

}
\newpage
\onecolumn
\section*{Supplementary Materials}
Tab. \ref{tab:orbital}--\ref{tab:notredame} present the results of the error criteria comparison from each individual dataset considered in the main paper.
See Tab. \ref{tab:datasets} for the statistics of the datasets.
Note that combining the results in Tab. \ref{tab:orbital}--\ref{tab:notredame} leads to Tab. 2 of the main paper.

{\renewcommand{\arraystretch}{1.2}%
\begin{table*}[ht]
\small
\begin{center}
\begin{tabular}{|l|rrr|rrrrr|}
\hline
& Orbital & Lateral & Forward & Dinosaur & Model House & Corridor & Fountain & Notre Dame \\
\hline
Cameras & 16 & 2 & 2 & 36 & 10 & 11 & 14 & 715\\
Points & $10^5$& $10^5$ & $10^5$ & 4,983 & 672 & 737 & 5,302 & 127,431 \\
Matches &$10^5$ & $10^5$& $10^5$& 27,080 & 5,550 & 12,139& 58,815 & 5,120,077\\
\hline
\end{tabular}
\end{center}
\caption{Datasets used. ``Matches" denotes the number of pairwise instances where a point is visible in two camera views. Generally, the number of matches is larger than the number of points because the points may be visible in more than two views.}
\label{tab:datasets}
\end{table*}
}\vspace{-0.5em}

{\renewcommand{\arraystretch}{1.2}%
\begin{table*}[ht]
\small
\begin{center}
\begin{tabular}{cl|cccccccc|}
\cline{3-10}
&& Midpoint & $L_1$ img & $L_2$ img& $L_2$ img & $L_\infty$ img & \multirow{2}{*}{$L_1$ ang} & \multirow{2}{*}{$L_2$ ang} & \multirow{2}{*}{$L_\infty$ ang}\\
 &  &\cite{midpoint, hartley_triangulation}&\cite{hartley_triangulation} & \cite{hartley_triangulation}& 5 it. \cite{lindstrom} & \cite{nister_phd_thesis} &&&\\
\hline
\multicolumn{1}{|c|}{\parbox[t]{2mm}{\multirow{7}{*}{\rotatebox[origin=c]{90}{Error Criterion}}}} & $\theta_0+\theta_1$  & - & -  & -  & -  & - & 100,000 & -  & -  \\
\multicolumn{1}{|c|}{}&$\theta_0^2+\theta_1^2$ & -  & -  & -  & -& - & - & 100,000  & -  \\
\multicolumn{1}{|c|}{}&$\sin^2(\theta_0)+\sin^2(\theta_1)$ & -  & -  & -  & -  & - & - & 100,000  & -  \\
\multicolumn{1}{|c|}{}&$\max(\theta_0, \theta_1)$  & -  & -  & -  & -  & - & - & -  & 100,000 \\
\cline{2-10}
\multicolumn{1}{|c|}{}&$d_0+d_1$  & - & 81,654& -& -& - & 18,346 & - & -\\
\multicolumn{1}{|c|}{}& $d_0^2+d_1^2$  & - & - & 23,435 & 76,565& -&- &- &- \\
\multicolumn{1}{|c|}{}& $\max(d_0, d_1)$  & -  & -  & -  & -  & 100,000 & - & -  & -  \\ 
\hline
\end{tabular}
\end{center}
\caption{\textbf{[Orbital]} The number of matches for which each method yields the lowest error in given criterion.}
\label{tab:orbital}
\end{table*}
}\vspace{-0.5em}

{\renewcommand{\arraystretch}{1.2}%
\begin{table*}[ht]
\small
\begin{center}
\begin{tabular}{cl|cccccccc|}
\cline{3-10}
&& Midpoint & $L_1$ img & $L_2$ img& $L_2$ img & $L_\infty$ img & \multirow{2}{*}{$L_1$ ang} & \multirow{2}{*}{$L_2$ ang} & \multirow{2}{*}{$L_\infty$ ang}\\
 &  &\cite{midpoint, hartley_triangulation}&\cite{hartley_triangulation} & \cite{hartley_triangulation}& 5 it. \cite{lindstrom} & \cite{nister_phd_thesis} &&&\\
\hline
\multicolumn{1}{|c|}{\parbox[t]{2mm}{\multirow{7}{*}{\rotatebox[origin=c]{90}{Error Criterion}}}} & $\theta_0+\theta_1$  & - & -  & -  & -  & - & 100,000 & -  & -  \\
\multicolumn{1}{|c|}{}&$\theta_0^2+\theta_1^2$ & -  & -  & -  & -& - & - & 100,000  & -  \\
\multicolumn{1}{|c|}{}&$\sin^2(\theta_0)+\sin^2(\theta_1)$ & -  & -  & -  & -  & - & - & 100,000  & -  \\
\multicolumn{1}{|c|}{}&$\max(\theta_0, \theta_1)$  & -  & -  & -  & -  & - & - & -  & 100,000 \\
\cline{2-10}
\multicolumn{1}{|c|}{}&$d_0+d_1$  & - & 99,814 & 1& 2& - & 183 & - & -\\
\multicolumn{1}{|c|}{}& $d_0^2+d_1^2$  & - & - & 11,485 & 88,515& -&- &- &- \\
\multicolumn{1}{|c|}{}& $\max(d_0, d_1)$  & -  & -  & -  & -  & 100,000 & - & -  & -  \\ 
\hline
\end{tabular}
\end{center}
\caption{\textbf{[Lateral]} The number of matches for which each method yields the lowest error in given criterion.}
\label{tab:lateral}
\end{table*}
}\vspace{-0.5em}

{\renewcommand{\arraystretch}{1.2}%
\begin{table*}[ht]
\small
\begin{center}
\begin{tabular}{cl|cccccccc|}
\cline{3-10}
&& Midpoint & $L_1$ img & $L_2$ img& $L_2$ img & $L_\infty$ img & \multirow{2}{*}{$L_1$ ang} & \multirow{2}{*}{$L_2$ ang} & \multirow{2}{*}{$L_\infty$ ang}\\
 &  &\cite{midpoint, hartley_triangulation}&\cite{hartley_triangulation} & \cite{hartley_triangulation}& 5 it. \cite{lindstrom} & \cite{nister_phd_thesis} &&&\\
\hline
\multicolumn{1}{|c|}{\parbox[t]{2mm}{\multirow{7}{*}{\rotatebox[origin=c]{90}{Error Criterion}}}} & $\theta_0+\theta_1$  & - & -  & -  & -  & - & 100,000 & -  & -  \\
\multicolumn{1}{|c|}{}&$\theta_0^2+\theta_1^2$ & -  & -  & 4  & 3 & - & - & 99,993  & -  \\
\multicolumn{1}{|c|}{}&$\sin^2(\theta_0)+\sin^2(\theta_1)$ & -  & -  & -  & -  & - & - & 100,000  & -  \\
\multicolumn{1}{|c|}{}&$\max(\theta_0, \theta_1)$  & -  & -  & -  & -  & - & - & -  & 100,000 \\
\cline{2-10}
\multicolumn{1}{|c|}{}&$d_0+d_1$  & - & 7,625& -&-& - & 92,375  & - & -\\
\multicolumn{1}{|c|}{}& $d_0^2+d_1^2$  & - & - & 34,785  & 65,215& -&- &- &- \\
\multicolumn{1}{|c|}{}& $\max(d_0, d_1)$  & -  & -  & -  & -  & 100,000 & - & -  & -  \\ 
\hline
\end{tabular}
\end{center}
\caption{\textbf{[Forward]} The number of matches for which each method yields the lowest error in given criterion.}
\label{tab:forward}
\end{table*}
}\vspace{-0.5em}

{\renewcommand{\arraystretch}{1.2}%
\begin{table*}[ht]
\small
\begin{center}
\begin{tabular}{cl|cccccccc|}
\cline{3-10}
&& Midpoint & $L_1$ img & $L_2$ img& $L_2$ img & $L_\infty$ img & \multirow{2}{*}{$L_1$ ang} & \multirow{2}{*}{$L_2$ ang} & \multirow{2}{*}{$L_\infty$ ang}\\
 &  &\cite{midpoint, hartley_triangulation}&\cite{hartley_triangulation} & \cite{hartley_triangulation}& 5 it. \cite{lindstrom} & \cite{nister_phd_thesis} &&&\\
\hline
\multicolumn{1}{|c|}{\parbox[t]{2mm}{\multirow{7}{*}{\rotatebox[origin=c]{90}{Error Criterion}}}} & $\theta_0+\theta_1$  & - & -  & -  & -  & - & 27,080 & -  & -  \\
\multicolumn{1}{|c|}{}&$\theta_0^2+\theta_1^2$ & -  & -  & -  & -& - & - & 27,080  & -  \\
\multicolumn{1}{|c|}{}&$\sin^2(\theta_0)+\sin^2(\theta_1)$ & -  & -  & -  & -  & - & - & 27,080  & -  \\
\multicolumn{1}{|c|}{}&$\max(\theta_0, \theta_1)$  & -  & -  & -  & -  & - & - & -  & 27,080 \\
\cline{2-10}
\multicolumn{1}{|c|}{}&$d_0+d_1$  & - & 27,080& -& -& - & - & - & -\\
\multicolumn{1}{|c|}{}& $d_0^2+d_1^2$  & - & - & 6,598 & 20,482 &-&- &- &- \\
\multicolumn{1}{|c|}{}& $\max(d_0, d_1)$  & -  & -  & -  & -  & 27,080 & - & -  & -  \\ 
\hline
\end{tabular}
\end{center}
\caption{\textbf{[Dinosaur]} The number of matches for which each method yields the lowest error in given criterion.}
\label{tab:dinosaur}
\end{table*}
}\vspace{-0.5em}

{\renewcommand{\arraystretch}{1.2}%
\begin{table*}[ht]
\small
\begin{center}
\begin{tabular}{cl|cccccccc|}
\cline{3-10}
&& Midpoint & $L_1$ img & $L_2$ img& $L_2$ img & $L_\infty$ img & \multirow{2}{*}{$L_1$ ang} & \multirow{2}{*}{$L_2$ ang} & \multirow{2}{*}{$L_\infty$ ang}\\
 &  &\cite{midpoint, hartley_triangulation}&\cite{hartley_triangulation} & \cite{hartley_triangulation}& 5 it. \cite{lindstrom} & \cite{nister_phd_thesis} &&&\\
\hline
\multicolumn{1}{|c|}{\parbox[t]{2mm}{\multirow{7}{*}{\rotatebox[origin=c]{90}{Error Criterion}}}} & $\theta_0+\theta_1$  & - & -  & -  & -  & - & 5,550 & -  & -  \\
\multicolumn{1}{|c|}{}&$\theta_0^2+\theta_1^2$ & -  & -  & -  & -& - & - & 5,550  & -  \\
\multicolumn{1}{|c|}{}&$\sin^2(\theta_0)+\sin^2(\theta_1)$ & -  & -  & -  & -  & - & - & 5,550  & -  \\
\multicolumn{1}{|c|}{}&$\max(\theta_0, \theta_1)$  & -  & -  & -  & -  & - & - & -  & 5,550 \\
\cline{2-10}
\multicolumn{1}{|c|}{}&$d_0+d_1$  & - & 3,563 & - & - & - & 1,987 & - & -\\
\multicolumn{1}{|c|}{}& $d_0^2+d_1^2$  & - & - & 2,766 & 2,784& -&- &- &- \\
\multicolumn{1}{|c|}{}& $\max(d_0, d_1)$  & -  & -  & -  & -  & 5,550 & - & -  & -  \\ 
\hline
\end{tabular}
\end{center}
\caption{\textbf{[Model House]} The number of matches for which each method yields the lowest error in given criterion.}
\label{tab:modelhouse}
\end{table*}
}\vspace{-0.5em}

{\renewcommand{\arraystretch}{1.2}%
\begin{table*}[ht]
\small
\begin{center}
\begin{tabular}{cl|cccccccc|}
\cline{3-10}
&& Midpoint & $L_1$ img & $L_2$ img& $L_2$ img & $L_\infty$ img & \multirow{2}{*}{$L_1$ ang} & \multirow{2}{*}{$L_2$ ang} & \multirow{2}{*}{$L_\infty$ ang}\\
 &  &\cite{midpoint, hartley_triangulation}&\cite{hartley_triangulation} & \cite{hartley_triangulation}& 5 it. \cite{lindstrom} & \cite{nister_phd_thesis} &&&\\
\hline
\multicolumn{1}{|c|}{\parbox[t]{2mm}{\multirow{7}{*}{\rotatebox[origin=c]{90}{Error Criterion}}}} & $\theta_0+\theta_1$  & - & -  & -  & -  & - & 12,139 & -  & -  \\
\multicolumn{1}{|c|}{}&$\theta_0^2+\theta_1^2$ & -  & -  & -  & - & - & - & 12,139  & -  \\
\multicolumn{1}{|c|}{}&$\sin^2(\theta_0)+\sin^2(\theta_1)$ & -  & -  & -  & -  & - & - & 12,139  & -  \\
\multicolumn{1}{|c|}{}&$\max(\theta_0, \theta_1)$  & -  & -  & -  & -  & - & - & -  & 12,139 \\
\cline{2-10}
\multicolumn{1}{|c|}{}&$d_0+d_1$  & - & 6,053& -&-& - & 6,086  & - & -\\
\multicolumn{1}{|c|}{}& $d_0^2+d_1^2$  & - & - & 5,999  & 6,140& -&- &- &- \\
\multicolumn{1}{|c|}{}& $\max(d_0, d_1)$  & -  & -  & -  & -  & 12,139 & - & -  & -  \\ 
\hline
\end{tabular}
\end{center}
\caption{\textbf{[Corridor]} The number of matches for which each method yields the lowest error in given criterion.}
\label{tab:corridor}
\end{table*}
}\vspace{-0.5em}
{\renewcommand{\arraystretch}{1.2}%
\begin{table*}[ht]
\small
\begin{center}
\begin{tabular}{cl|cccccccc|}
\cline{3-10}
&& Midpoint & $L_1$ img & $L_2$ img& $L_2$ img & $L_\infty$ img & \multirow{2}{*}{$L_1$ ang} & \multirow{2}{*}{$L_2$ ang} & \multirow{2}{*}{$L_\infty$ ang}\\
 &  &\cite{midpoint, hartley_triangulation}&\cite{hartley_triangulation} & \cite{hartley_triangulation}& 5 it. \cite{lindstrom} & \cite{nister_phd_thesis} &&&\\
\hline
\multicolumn{1}{|c|}{\parbox[t]{2mm}{\multirow{7}{*}{\rotatebox[origin=c]{90}{Error Criterion}}}} & $\theta_0+\theta_1$  & - & -  & -  & -  & - & 58,815 & -  & -  \\
\multicolumn{1}{|c|}{}&$\theta_0^2+\theta_1^2$ & -  & -  & -  & - & - & - & 58,815  & -  \\
\multicolumn{1}{|c|}{}&$\sin^2(\theta_0)+\sin^2(\theta_1)$ & -  & -  & -  & -  & - & - & 58,815  & -  \\
\multicolumn{1}{|c|}{}&$\max(\theta_0, \theta_1)$  & -  & -  & -  & -  & - & - & -  & 58,815 \\
\cline{2-10}
\multicolumn{1}{|c|}{}&$d_0+d_1$  & - & 32,557& -&-& - & 26,258  & - & -\\
\multicolumn{1}{|c|}{}& $d_0^2+d_1^2$  & - & - & 2,348  & 56,467& -&- &- &- \\
\multicolumn{1}{|c|}{}& $\max(d_0, d_1)$  & -  & -  & -  & -  & 58,815 & - & -  & -  \\ 
\hline
\end{tabular}
\end{center}
\caption{\textbf{[Fountain]} The number of matches for which each method yields the lowest error in given criterion.}
\label{tab:fountain}
\end{table*}
}\vspace{-0.5em}

{\renewcommand{\arraystretch}{1.2}%
\begin{table*}[t]
\small
\begin{center}
\begin{tabular}{cl|cccccccc|}
\cline{3-10}
&& Midpoint & $L_1$ img & $L_2$ img& $L_2$ img & $L_\infty$ img & \multirow{2}{*}{$L_1$ ang} & \multirow{2}{*}{$L_2$ ang} & \multirow{2}{*}{$L_\infty$ ang}\\
 &  &\cite{midpoint, hartley_triangulation}&\cite{hartley_triangulation} & \cite{hartley_triangulation}& 5 it. \cite{lindstrom} & \cite{nister_phd_thesis} &&&\\
\hline
\multicolumn{1}{|c|}{\parbox[t]{2mm}{\multirow{7}{*}{\rotatebox[origin=c]{90}{Error Criterion}}}} & $\theta_0+\theta_1$  & - & -  & -  & -  & - & 5,120,077 & -  & -  \\
\multicolumn{1}{|c|}{}&$\theta_0^2+\theta_1^2$ & -  & -  & -  & - & - & - & 5,120,077  & -  \\
\multicolumn{1}{|c|}{}&$\sin^2(\theta_0)+\sin^2(\theta_1)$ & -  & -  & -  & -  & - & - & 5,120,077  & -  \\
\multicolumn{1}{|c|}{}&$\max(\theta_0, \theta_1)$  & -  & -  & -  & -  & - & - & -  & 5,120,077 \\
\cline{2-10}
\multicolumn{1}{|c|}{}&$d_0+d_1$  & - & 3,654,620 & 121 & 123& - & 1,465,213  & - & -\\
\multicolumn{1}{|c|}{}& $d_0^2+d_1^2$  & - & - & 1,190,827  & 3,929,250 & -&- &- &- \\
\multicolumn{1}{|c|}{}& $\max(d_0, d_1)$  & -  & -  & -  & -  & 5,120,077 & - & -  & -  \\ 
\hline
\end{tabular}
\end{center}
\caption{\textbf{[Notre Dame]} The number of matches for which each method yields the lowest error in given criterion.}
\vspace*{17.5cm}
\label{tab:notredame}
\end{table*}
}

\end{document}